\documentclass[10pt,twocolumn,letterpaper]{article}

\usepackage[pagenumbers]{cvpr} 

\makeatletter
\@namedef{ver@everyshi.sty}{}
\makeatother
\usepackage{graphicx}
\usepackage{amsmath}
\usepackage{amssymb}
\usepackage{booktabs}
\usepackage{multirow}
\usepackage{bbm}
\usepackage{color}
\usepackage{colortbl}
\definecolor{backcolor}{RGB}{232, 242, 255}
\usepackage[accsupp]{axessibility}  

\newlength\savewidth\newcommand\shline{\noalign{\global\savewidth\arrayrulewidth
  \global\arrayrulewidth 1pt}\hline\noalign{\global\arrayrulewidth\savewidth}}
\newcommand{\tablestyle}[2]{\setlength{\tabcolsep}{#1}\renewcommand{\arraystretch}{#2}\centering\footnotesize}
\newcommand\blfootnote[1]{\begingroup\renewcommand\thefootnote{}\footnote{#1}\addtocounter{footnote}{-1}\endgroup}

\newcolumntype{x}[1]{>{\centering\arraybackslash}p{#1pt}}
\newcolumntype{y}[1]{>{\raggedright\arraybackslash}p{#1pt}}
\newcolumntype{z}[1]{>{\raggedleft\arraybackslash}p{#1pt}}

\newcommand{\app}{\raise.17ex\hbox{$\scriptstyle\sim$}}

\definecolor{deemph}{gray}{0.6}

\definecolor{baselinecolor}{gray}{.9}
\newcommand{\baseline}[1]{\cellcolor{baselinecolor}{#1}}

\definecolor{G}{RGB}{112,173,71}
\definecolor{O}{RGB}{237,125,49}

\usepackage{pgffor}

\usepackage[pagebackref,breaklinks,colorlinks]{hyperref}

\usepackage[capitalize]{cleveref}
\crefname{section}{Sec.}{Secs.}
\Crefname{section}{Section}{Sections}
\Crefname{table}{Table}{Tables}
\crefname{table}{Tab.}{Tabs.}

\def\cvprPaperID{***} 
\def\confName{CVPR}
\def\confYear{2023}

\begin{document}

\title{Mixed Autoencoder for Self-supervised Visual Representation Learning}

\author{
    Kai Chen$^{1*}$
    \quad
    Zhili Liu$^{1,2*}$
    \quad
    Lanqing Hong$^{2}$
    \quad
    Hang Xu$^{2}$
    \quad
    Zhenguo Li$^{2}$
    \quad
    Dit-Yan Yeung$^{1}$
    \\
    $^{1}$Hong Kong University of Science and Technology
    \quad
    $^2$Huawei Noah's Ark Lab
    \\
    {\tt\small \{kai.chen,zhili.liu\}@connect.ust.hk
    \quad
    \{honglanqing,xu.hang,li.zhenguo\}@huawei.com} 
    \\
    {\tt\small dyyeung@cse.ust.hk}
}

\maketitle
\vspace{-6mm}


\blfootnote{
$^{*}$Equal contribution.
}
\begin{abstract}

   Masked Autoencoder (MAE) has demonstrated superior performance on various vision tasks via randomly masking image patches and reconstruction.
   However, effective data augmentation strategies for MAE still remain open questions, different from those in contrastive learning that serve as the most important part. 
   This paper studies the prevailing mixing augmentation for MAE.
   We first demonstrate that na\"ive mixing will in contrast degenerate model performance due to the increase of mutual information (MI).
   To address, we propose homologous recognition, an auxiliary pretext task, not only to alleviate the MI increasement by explicitly requiring each patch to recognize homologous patches, 
   but also to perform object-aware self-supervised pre-training for better downstream dense perception performance.
   With extensive experiments, 
   we demonstrate that our proposed Mixed Autoencoder (\textit{MixedAE}) achieves the state-of-the-art transfer results among masked image modeling (MIM) augmentations on different downstream tasks with significant 
   efficiency.
   Specifically, our \textit{MixedAE} outperforms MAE by 
   \textbf{+0.3\% accuracy}, \textbf{+1.7 mIoU} and \textbf{+0.9 AP} on ImageNet-1K, ADE20K and COCO respectively with a standard ViT-Base. 
   Moreover, \textit{MixedAE} surpasses iBOT, a strong MIM method combined with instance discrimination, while accelerating training by 2$\times$.
   To our best knowledge, this is the very first work to consider mixing for MIM from the perspective of pretext task design.
   Code will be made available.

\end{abstract}


\vspace{-6mm}
\section{Introduction}\label{sec:intro}


\begin{figure}[!t]
  \centering
   \includegraphics[width=1.0\linewidth]{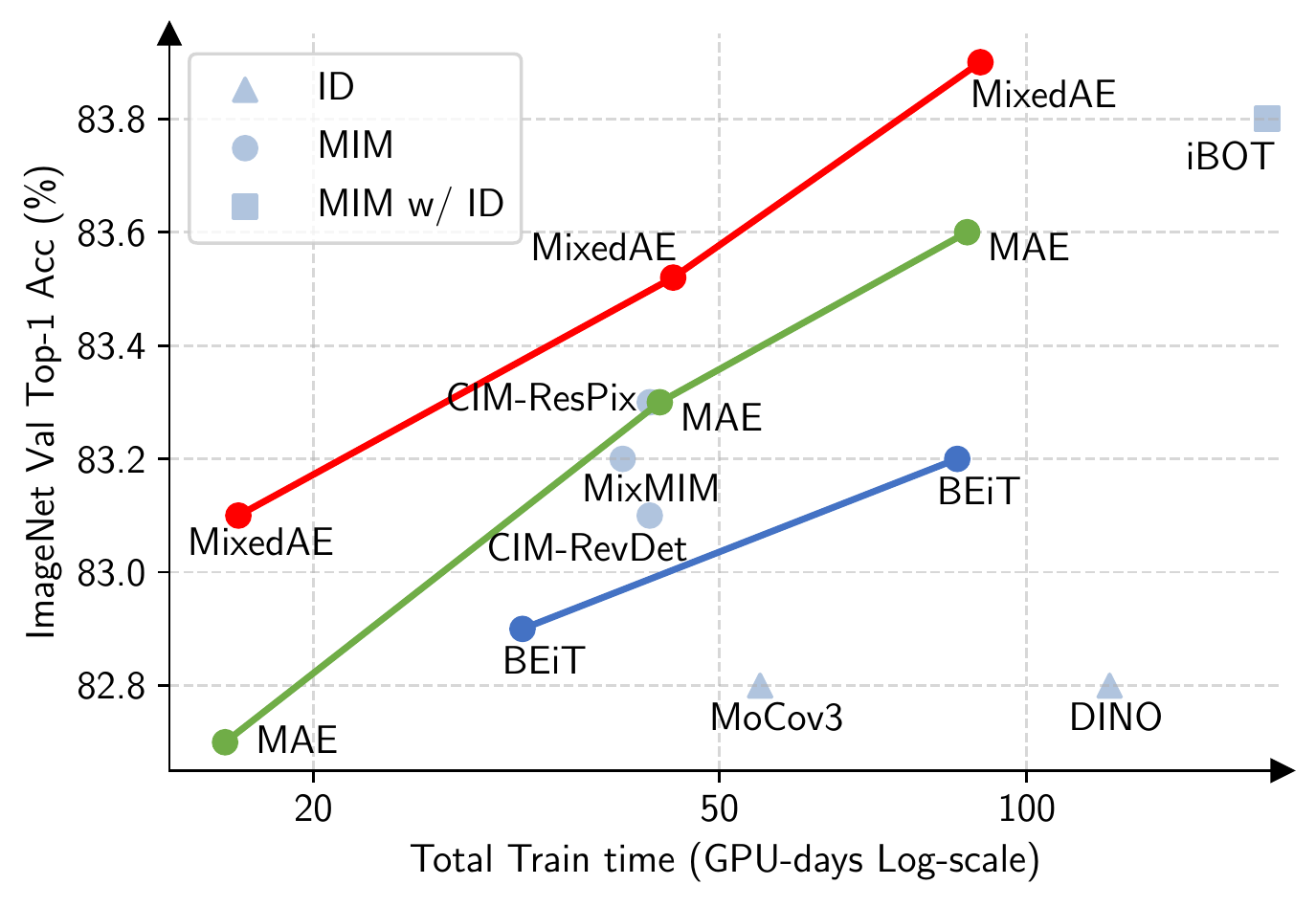}
   \vspace{-7mm}
   \caption{\textbf{Fine-tuning accuracy on ImageNet-1K.}
   Our \textit{MixedAE} achieves the best trade-off between pre-training overhead and transfer performance.
   Specifically, \textit{MixedAE} surpasses MAE~\cite{MAE} consistently with only 3\% extra overhead, while outperforms the strong iBOT~\cite{ibot} with only 53.4\% of its computation overhead.
   See more detailed comparisons in~\cref{tab:main_transfer}.
   ID stands for instance discrimination, while MIM represents masked image modeling.
   }
   \vspace{-5mm}
   \label{fig:tradeoff}
\end{figure}


Self-supervised learning (SSL) has become one of the most popular pre-training paradigm due to its independence of human annotation.
Previous literature mainly focuses on the handcrafted pretext task design~\cite{doersch2015unsupervised,pathak2016context,gidaris2018unsupervised} and instance discrimination~\cite{MOCOV3,DINO}, while with the development of Vision Transformer~\cite{VIT}, masked image modeling (MIM), deeply motivated by masked language modeling~\cite{devlin2018bert}, has started to demonstrate more superior effectiveness by firstly \textbf{masking} some patches of the input images and then \textbf{reconstructing} the 
masked patches from visible ones by predicting certain targets generated by masked patches.
In order to complete reconstruction, the encoder is expected to generate highly semantic representation which can be better transferred to downstream tasks~\cite{ADE20K,lin2014microsoft,han2021soda10m,li2022coda}
for superior performance.

Existing MIM works mainly concentrate on the design of the reconstruction targets (\eg, visual tokenizers~\cite{BEIT,PeCo}, pixels~\cite{MAE,SimMIM}, graphical features~\cite{MaskFeat} and instance discrimination~\cite{el2021large,ibot,data2vec}) and masking strategies (\eg, random~\cite{BEIT,MAE}, attention-guide~\cite{kakogeorgiou2022hide} and sample-dependent~\cite{shi2022adversarial}). 
See more detailed discussions in~\cref{sec:related}.
Despite the superior performance, 
we observe that the input augmentations for MIM have been seldom explored.
Specifically, adding \textit{color jittering}, an essential augmentation technique of contrastive learning~\cite{chen2020simple}, with MAE~\cite{MAE} even \textit{degrades} transfer results, suggesting that MIM might possess a different preference for data augmentations, and the effective data augmentation strategies for MIM are still an open question.

In this paper, we explore the usage of image mixing, a commonly used technique in both supervised~\cite{Mixup,CutMix} and contrastive learning~\cite{shen2022unmix,xu2022seed}, with MAE~\cite{MAE}.
We start by constructing a simple baseline to adopt mixing with MAE directly, which, different from in supervised and contrastive learning, would instead ease the reconstruction pretext by increasing the \textit{mutual information} between the model input and reconstruction target due to the usage of image mixing with global self-attention as proved in~\cref{sec:baseline}.
To address this issue, we propose \textit{homologous recognition}, an auxiliary pretext task to enforce each patch to recognize homologous patches explicitly according to attention distributions before reconstruction, and build our Mixed Autoencoder network (\textit{MixedAE}) in~\cref{sec:loss}.
Moreover, we demonstrate that our simple yet effective method can not only achieve significant performance improvement, but also conduct object-aware SSL pre-training without any specifically designed modules for better downstream dense perception results in~\cref{sec:discussion}.

Concurrently, MixMIM~\cite{liu2022mixmim} also considers mixing with MAE, but different from ours, 
1) \textbf{Purpose}: MixMIM uses mixing to recover the 2D structure after random masking for an efficient implementation to conduct MAE-style pre-training on hierarchy Vision Transformers~\cite{liu2021swin}, while ours utilizes mixing to conduct object-aware SSL pre-training for better representation learning.
2) \textbf{Method}: MixMIM uses masked self-attention to only perform attention within patches from the same images given the mixing masks as \textbf{input}, sharing the exactly same pretext task with MAE, while ours requires explicit homologous recognition given mixing masks as \textbf{target}, actively emerging mixing into the pretext design.
3) \textbf{Formulation}: The mixing ratio $r$ is limited to 0.5 in MixMIM, which instead can be flexibly selected from $(0, 0.5]$ in our formulation. See more details in~\cref{sec:method}.

The main contributions of this work contain three parts:
\begin{enumerate}
    \item We propose the Mixed Autoencoder (\textit{MixedAE}), a simple yet effective approach to conduct object-aware pre-training without introducing any specifically designed modules.
    With extensive experiments, we demonstrate that \textit{MixedAE} can achieve the state-of-the-art transfer performance on various downstream tasks including image classification, semantic segmentation and object detection, while maintaining significant efficiency.
    
    \item We theoretically demonstrate the underlying design differences between MIM and previous supervision with mixing (\eg, supervised and contrastive learning).
    
    \item  To our best knowledge, this is the first work to consider mixing as an effective data augmentation strategy for MIM from the perspective of pretext design with a pure autoencoder-based architecture.
    
\end{enumerate}



\begin{figure*}[t]
  \centering
  \includegraphics[width=0.9\linewidth]{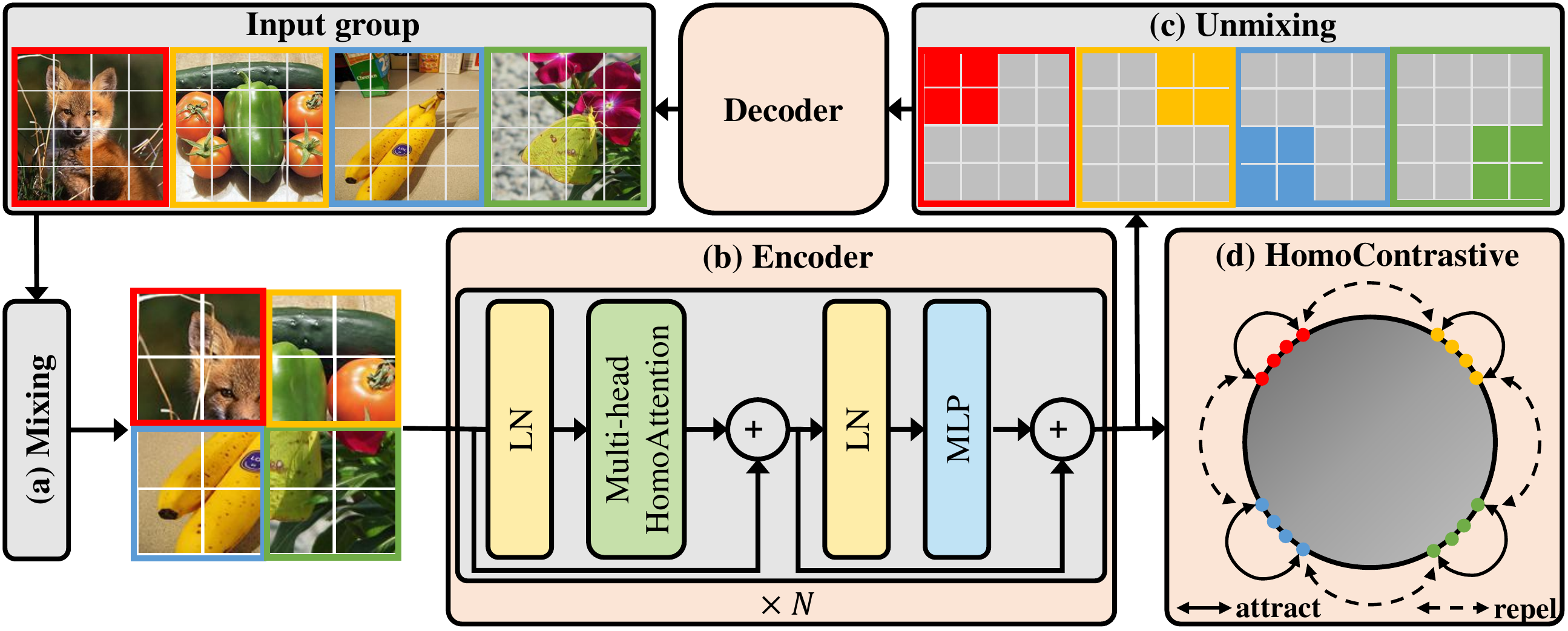}
  \vspace{-1mm}
   \caption{\textbf{Model architecture of Mixed Autoencoder (\textit{MixedAE})}.
   (a) The input images are first separated into groups to generate mixed samples independently, which are further taken as input to the encoder for feature extraction.
   (b) The self-attention operations are replaced with our homologous attention, enforcing each patch to only attend to patches with the highest attention mass.
   (c) The encoder features will be ``unmixed'' and fed into the decoder for pixel reconstruction.
   (d) Meanwhile, the homologous contrastive loss is adopted to verify the sampling accuracy by encouraging features of homologous patches to be similar, while heterologous ones to be dissimilar.
   }
  \vspace{-3mm}
   \label{fig:architecture}
\end{figure*}


\vspace{-2mm}
\section{Related Work}\label{sec:related}
Self-supervised learning aims at learning a transferable representation without human annotation.
Previous works mainly focus on handcrafted pretext design~\cite{doersch2015unsupervised,pathak2016context,gidaris2018unsupervised} and instance discrimination~\cite{chen2020simple,chen2020improved,grill2020bootstrap}.
Mask image modeling (MIM), inspired by the mask language modeling~\cite{devlin2018bert}, has achieved significant performance with superior pre-training efficiency by firstly masking portion of an image, and then reconstructing the masked part based on the visible one.

\vspace{-3.5mm}
\paragraph{Reconstruction target.}
BEiT~\cite{BEIT} pioneeringly proposes to predict visual tokens generated by a pre-trained visual tokenizer~\cite{ramesh2021zero}, which is simplified by SimMIM~\cite{SimMIM} to use pixel values as the reconstruction target directly.
MAE~\cite{MAE} proposes an asymmetric encoder-decoder architecture for better efficiency.
Besides pixels as the target, MaskFeat~\cite{MaskFeat} utilizes HOG features, while PeCo~\cite{PeCo} enhances the BEiT tokenizer with an additional perceptual loss.
Recent works combine the idea of instance discrimination~\cite{chen2020simple} with MIM.
iBOT~\cite{ibot} considers MIM as a self-distillation process with a Siamese architecture, and data2vec~\cite{data2vec} proposes a unified framework to conduct masked reconstruction pre-training for speech, images and languages.
SplitMask~\cite{el2021large} divides an image into two equal partitions and performs contrastive learning and MIM in the multi-task manner.
In this paper, we build \textit{MixedAE} based on MAE due to its efficiency and effectiveness, while the improvement brought by \textit{MixedAE} is complementary to more advanced reconstruction targets.

\vspace{-3.5mm}
\paragraph{Masking strategy.}
Instead of random masking~\cite{BEIT,MAE}, AttMask~\cite{kakogeorgiou2022hide} proposes a novel attention-guided masking strategy by masking according to the attention map of the final Transformer layer, while ADIOS~\cite{shi2022adversarial} introduces an adversarial objective between masking and reconstruction to generate learnable masks for MIM pre-training.
In this paper, we utilize random masking for \textit{MixedAE} following MAE due to its simplicity and effectiveness.

\vspace{-3.5mm}
\paragraph{Input augmentation} instead has been less explored for MIM. 
Instead of masking, CIM~\cite{fang2022corrupted} adopts a small BEiT as the generator to corrupt an image, which is further taken as input to an enhancer to reconstruct the corrupted patches  
or distinguish the corrupted patches from the uncorrupted ones.
Concurrently, MixMIM~\cite{liu2022mixmim} considers mixing with MAE also, but different from ours, MixMIM uses \textit{masked self-attention} to only perform attention within patches from the same images given the mixing masks as \textit{input}, 
which is exactly the same with MAE from the perspective of pretext design, 
while ours utilizes mixing as part of the pretext task actively to conduct object-aware SSL pre-training.


\section{Method}\label{sec:method}

In this section, we start by adopting mixing in MAE~\cite{MAE} with a simple baseline in~\cref{sec:baseline}, which, as we can prove, will instead ease the reconstruction pretext task.
Then, we propose a novel auxiliary pretext task to formulate our final \textit{MixedAE}, which can not only alleviate the ease of reconstruction, but also achieve object-aware SSL pre-training without specifically designed modules in~\cref{sec:homo,sec:loss,sec:discussion}. 


\subsection{Mixing: A Simple Baseline}\label{sec:baseline}

Given a unlabeled dataset, we randomly sample a \textit{clean} data batch with size $B$, which are later divided into non-overlapping patch sequences $\{\boldsymbol{x}_i\}_{i=1}^B$ $(\boldsymbol{x}_i\in\mathcal{R}^{L\times(P^2\cdot C)})$ following ViT~\cite{VIT}, where $L$ is the sequence length, $P$ is the patch size, and $C$ is the image channel dimension.


\paragraph{Mixing.}
The data batch is further separated into \textit{groups} $\{\{\boldsymbol{x}_i^j\}_{i=1}^{1/r}\}_{j=1}^{Br}$, and each group will generate a single \textit{mixed} image, where $r\in(0, 0.5]$ is the \textit{mixing ratio} representing the ratio of patches each clean image contributes to within a single mixed sample.
Different from MixMIM~\cite{liu2022mixmim}, $r$ is not restricted to 0.5 in our formulation.
Therefore, the mixing process for the $j$-th group can be represented as,
\begin{equation}
    \boldsymbol{\hat{x}}^j = \sigma_{mix}(\{\boldsymbol{x}_i^j\}, \boldsymbol{M}^j) = \sum_{i=1}^{1/r}\mathbbm{1}(\boldsymbol{M}^j = i)\boldsymbol{x}_i^j,
    \label{equ:mixing}
\end{equation}
where $\mathbbm{1}(\cdot)$ is indicator function and
$\boldsymbol{M}^j\in\{1,2,...,1/r\}^L$ represents a random mixing mask independently generated for the $j$-th group, which satisfies,
\begin{equation}
    \sum_{l=1}^L\mathbbm{1}(M^j_l = i) = rL,\ \forall i\in\{ 1,2,...,1/r\}.
\end{equation}
So, $\boldsymbol{M}^j$ determines the source patch in each position of $\boldsymbol{\hat{x}}^j$, while maintaining the mixing ratio for each clean image equal with $r$ (\ie symmetric mixing).
The mixed images $\boldsymbol{\hat{x}}^j$ are further fed into the encoder for feature extraction, which can be represented as 
$\boldsymbol{\hat{z}}^j = \textsc{enc}(\boldsymbol{\hat{x}}^j)$.

\paragraph{Unmixing.}
Following MAE~\cite{MAE}, $\boldsymbol{\hat{z}}^j$ is then ``unmixed'' to recover the input batch before mixing by inserting a special \texttt{[MASK]} token with $\boldsymbol{M}^j$. For $\forall i\in\{1,2...,1/r\}$, we have,
\begin{equation}
    \boldsymbol{z}^j_i = \mathbbm{1}(\boldsymbol{M}^j = i)\boldsymbol{\hat{z}}^j + [1-\mathbbm{1}(\boldsymbol{M}^j = i)]\texttt{[MASK]}.
\end{equation}
The ``unmixed'' group $\{\boldsymbol{z}_i^j\}_{i=1}^{1/r}$ is then taken as input to the decoder for pixel reconstruction, as $\boldsymbol{y}_i^j = \textsc{dec}(\boldsymbol{z}_i^j)$. 
Finally, the reconstruction loss can be formulated as,
\begin{equation}
    \label{equ:reconstruction}
    \mathcal{L}_{recon} = \sum_{i=1}^{1/r}\sum_{l=1}^L[1-\mathbbm{1}(M^j_l = i)](\boldsymbol{y}_{i,l}^j - \boldsymbol{x}_{i,l}^j)^2.
\end{equation}

So far, we have built a simple baseline to adopt mixing for MAE, which, however, performs even worse than MAE, as demonstrated in~\cref{tab:homo_summary}.
In the following, we provide a theoretical explanation to prove that this na\"ive incorporation will actually \textbf{ease} the reconstruction pretext task.


\vspace{-2mm}
\paragraph{Mutual information analysis.}
Without loss of generality, we take $r=0.5$ as an example. 
Denote $\boldsymbol{X}_1,\boldsymbol{X}_2$ as two random variables representing two input images, and $\boldsymbol{X}_1$ is further considered as the reconstruction target (symmetric for $\boldsymbol{X}_2$).
Then, we can prove that the mutual information (MI) between the mixed input $\sigma_{mix}(\{\boldsymbol{X}_1,\boldsymbol{X}_2\}, \boldsymbol{M})$ and the target $\boldsymbol{X}_1$ is no smaller than that between the MAE input $\sigma_{MAE}(\boldsymbol{X}_1, \boldsymbol{M})$ and $\boldsymbol{X}_1$ as 
(see proofs in~\cref{app:mi_proof}),
\begin{equation}
    I(\sigma_{mix}(\{\boldsymbol{X}_1,\boldsymbol{X}_2\}, \boldsymbol{M});\boldsymbol{X}_1) \ge I(\sigma_{MAE}(\boldsymbol{X}_1, \boldsymbol{M});\boldsymbol{X}_1).
    \label{equ:mi}
\end{equation}
Therefore, different from masking, which is introduced to \textbf{decrease} the mutual information between the model input and the reconstruction target due to the redundancy of image signals~\cite{MAE}, na\"ive mixing will instead \textbf{increase} the MI, and thus, ease the reconstruction pretext task.
Verification experiment is conducted in~\cref{app:exp}.

Also note that the MI increasement brought by mixing is target-invariant, suggesting that~\cref{equ:mi} also holds when the target is \textit{semantic labels} for supervised learning or \textit{positive samples} for contrastive learning, for which MI increasement is appealing.
This might explain why na\"ive mixing without auxiliary supervision is beneficial for supervised~\cite{Mixup,CutMix} and contrastive learning~\cite{shen2022unmix,xu2022seed}, but not for MAE.



\begin{figure}[!t]
  \centering
  \begin{subfigure}{1.0\linewidth}
    \includegraphics[width=1.0\linewidth]{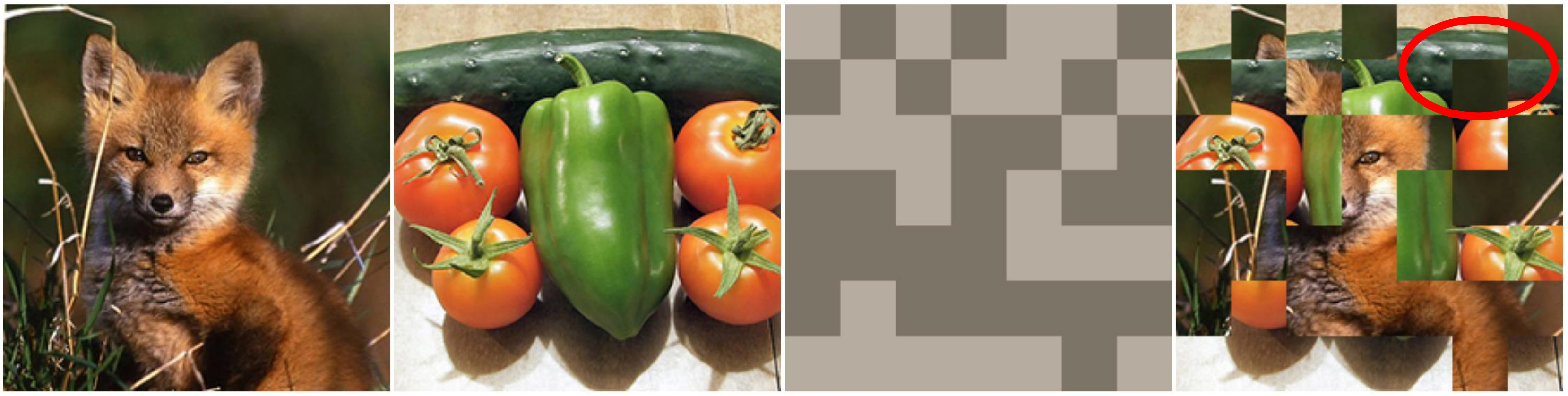}
    \caption{Uncertainty of generative modeling.}
    \label{fig:uncertainty}
  \end{subfigure}
  \begin{subfigure}{1.0\linewidth}
    \includegraphics[width=1.0\linewidth]{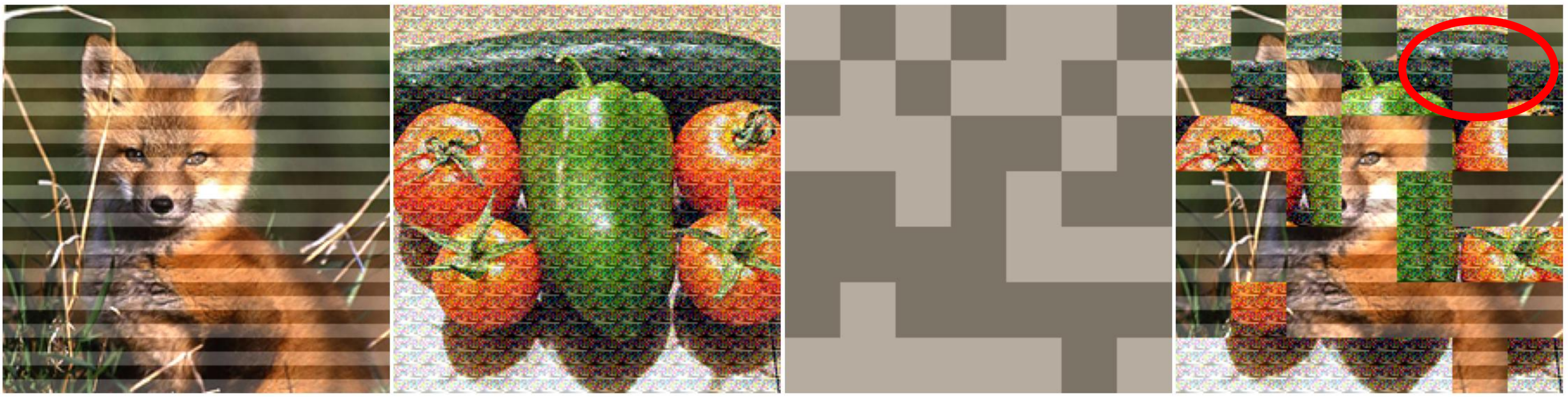}
    \caption{Different segment embeddings represent different images.}
    \label{fig:seg_embed}
  \end{subfigure}
  \caption{\textbf{Visualization of segment embeddings}.
  (a) Due to the uncertainty of generative modeling, green colors of the cucumber and the forest are both reasonable for patches in the red ellipse.
  (b) We adopt different segment embeddings for different images to provide necessary information for homologous recognition.
  }
  \label{fig:visual_seg_embed}
\end{figure}


\subsection{Recognition: Homologous Recognition}\label{sec:homo}

Another indispensable factor to achieve MI increasement is the usage of \textit{global self-attention} in ViT, with which each query patch will inevitably attend to heterologous patches from other images.
Due to the uncertainty of generative modeling, heterologous patches might provide a shortcut to complete reconstruction (\eg, the green color of cucumbers is a ``valuable'' cue to reconstruct the forest behind the fox in~\cref{fig:uncertainty}).
To address, we propose a novel auxiliary pretext task called \textit{homologous recognition} to enforce each query to explicitly recognize and only attend to homologous patches.

\vspace{-3mm}
\paragraph{Homologous attention} recognizes homologous patches on-the-fly by enforcing each query patch to only attend to key patches with the highest attention mass using a $\operatorname{TopK}(\cdot)$ sampling operation.
Specifically, the homologous attention can be formulated as,
\begin{equation}
    A_{HomoAtt} = \operatorname{softmax}(\operatorname{TopK}(\boldsymbol{q}\boldsymbol{k}^T/\sqrt{D_h})),
    \label{equ:homo_att}
\end{equation}
where $\boldsymbol{q}$ is the query patch, $\boldsymbol{k}$ are the key patches and $D_h$ is the feature dimension.
By default, all the self-attention operations in ViT are replaced with homologous attention except the very first layer. See comparisons in~\cref{tab:attn_layer}. 

\paragraph{Homologous contrastive} aims at verifying the $\operatorname{TopK}(\cdot)$ sampling accuracy by encouraging the encoder features of homologous patches to be similar, while heterologous ones to be dissimilar in the supervised contrastive manner~\cite{khosla2020supervised}.
The homologous contrastive loss can be formulated as,
\begin{equation}
    \mathcal{L}_{HomoCon} = -\sum_{l=1}^L\sum_{l^+}\log\frac{exp(cos(\boldsymbol{\hat{z}}^j_l, \boldsymbol{\hat{z}}^j_{l^+})/\tau)}{\sum_{l^\prime\neq l}^L exp(cos(\boldsymbol{\hat{z}}^j_l, \boldsymbol{\hat{z}}^j_{l^\prime})/\tau)},
    \label{equ:homo_con}
\end{equation}
where $\tau$ is the temperature and $cos(\cdot, \cdot)$ is the cosine similarity.
As demonstrated in~\cref{fig:accuracy}, the $\operatorname{TopK}(\cdot)$ sampling accuracy is significantly improved and stabilized with the usage of the homologous contrastive loss $\mathcal{L}_{HomoCon}$.

\vspace{-2mm}
\paragraph{Segment embedding.}
Beside the positional embeddings, we add another segment embedding to the mixed sequence $\boldsymbol{\hat{x}}^j$ following BERT~\cite{devlin2018bert} to provide necessary information to complete homologous recognition, due to the uncertainty of generative modeling.
The segment embedding is shared for patches from the same image, while different for patches from different images, as demonstrated in~\cref{fig:seg_embed}.



\begin{figure}[!t]
  \centering
  \begin{subfigure}{1.0\linewidth}
    \includegraphics[width=1.0\linewidth]{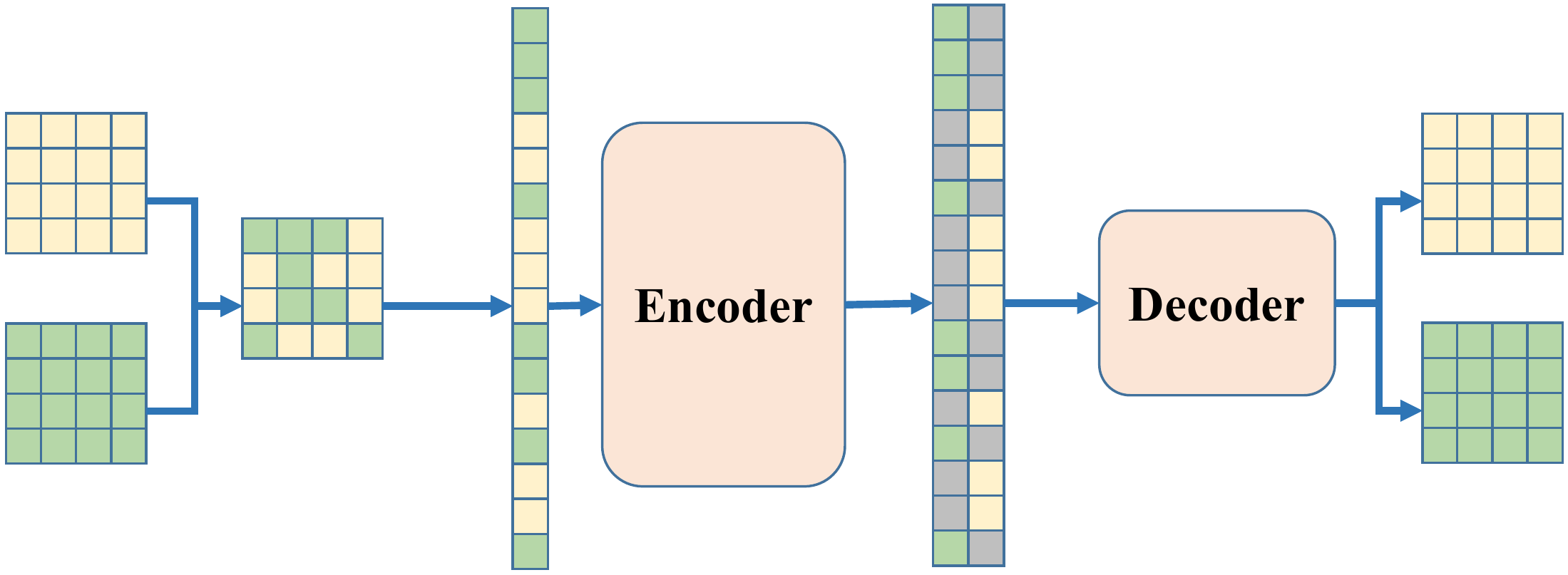}
    \caption{Compose mixing mode.}
    \label{fig:compose}
  \end{subfigure}
  \begin{subfigure}{1.0\linewidth}
    \includegraphics[width=1.0\linewidth]{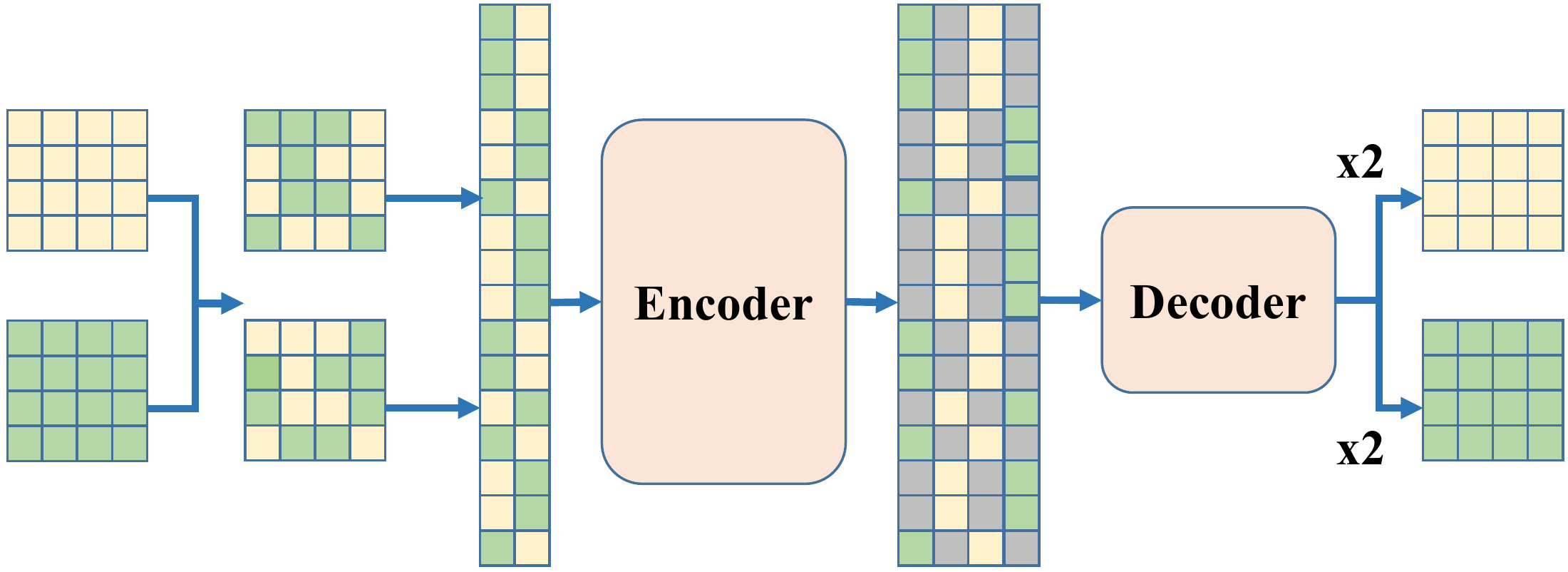}
    \caption{Full mixing mode.}
    \label{fig:full}
  \end{subfigure}
  \vspace{-6mm}
  \caption{\textbf{Visualization of two mixing modes when $r=0.5$}.
  (a) Each group generates a single mixed sample for the compose mixing mode, (b) while $1/r$ mixed samples are generated for the full mixing mode to maintain the effective batch size unchanged.}
  \vspace{-2mm}
  \label{fig:mixing}
\end{figure}


\vspace{-2mm}
\paragraph{Mixing mode.}
For a fair comparison under different training overheads, two mixing modes are adopted for \textit{MixedAE}, 1) \textbf{Compose}: each group generates a single mixed sample following~\cref{equ:mixing}, and the effective encoder batch size is $Br$; 2) \textbf{Full}: each group generates $1/r$ mixed samples by sampling $\boldsymbol{M}^j$ for $1/r$ times independently, and the effective encoder batch size is $B$.
An example is provided in~\cref{fig:mixing} when $r=0.5$.
As shown in~\cref{tab:main_transfer}, \textit{MixedAE} achieves the SoTA performance under different training overheads.
If no otherwise specified, compose mixing is adopted by default.



\begin{figure*}[!t]
    \centering
    \includegraphics[width=1.0\linewidth]{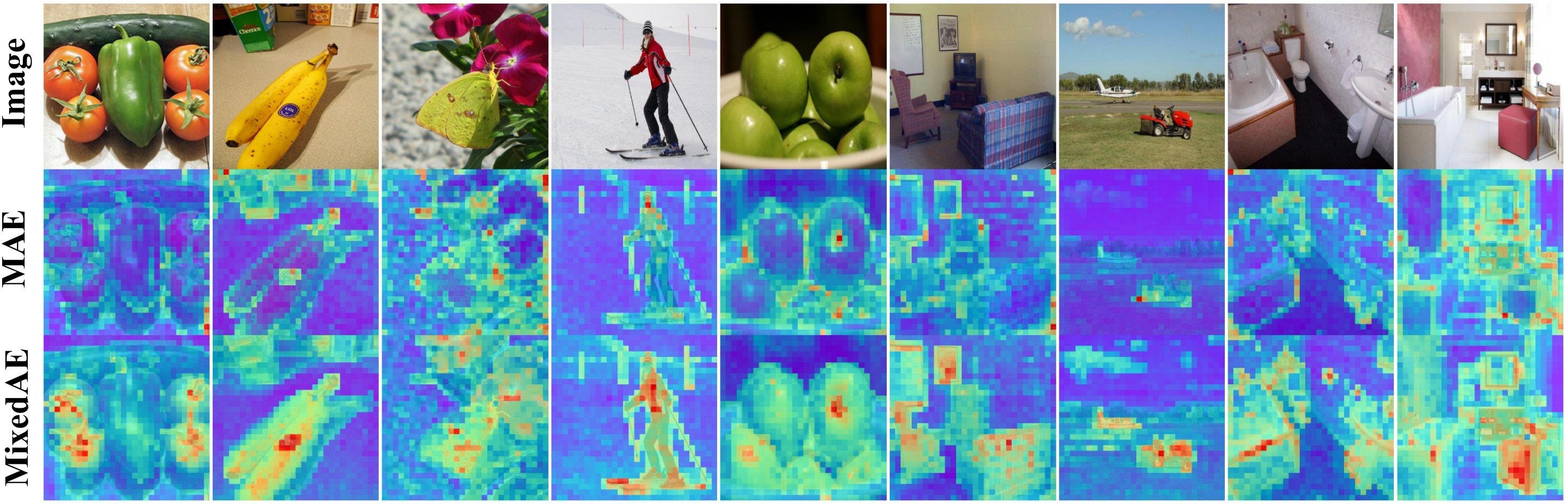}
    \vspace{-3mm}
    \caption{\textbf{Visualizations of attention maps} on images from ImageNet-1K~\cite{deng2009imagenet} (1st-3rd columns), Microsoft COCO~\cite{lin2014microsoft} (4th-6th columns) and ADE20K~\cite{ADE20K} datasets (7th-9th columns).
    Both MAE and \textit{MixedAE} are pre-trained on ImageNet-1K for 300 epochs. 
    Compared with MAE which mainly focuses on the most discriminative patches, (\eg, boundaries (1st, 2nd \& 5th) and edges (6th \& 8th)), \textit{MixedAE} discovers foreground object patches more precisely (3rd \& 9th) and completely (4th \& 7th).
    See more attention maps in~\cref{app:visualization}.
    }
    \vspace{-3mm}
    \label{fig:attention_map}
\end{figure*}


\subsection{Reconstruction: Loss Function}\label{sec:loss}

\paragraph{Loss function.}
We formulate \textit{MixedAE} in the multi-task learning manner and the final loss function is a weighted sum of the reconstruction loss $\mathcal{L}_{recon}$ and the homologous contrastive loss $\mathcal{L}_{HomoCon}$ as,
\begin{equation}
    \mathcal{L}_{MixedAE} = \mathcal{L}_{recon} + \lambda\mathcal{L}_{HomoCon},
\end{equation}
where the balanced weight $\lambda$ is set to be 0.1 by default.


\subsection{Discussion: Object-aware Pre-training}\label{sec:discussion}

With the usage of mixing, we observe that \textit{MixedAE} can achieve object-aware self-supervised pre-training without any specifically designed components, such as K-means~\cite{chen2021multisiam}, selective search~\cite{wei2021aligning} and object discovery network~\cite{henaff2022object}, because homologous recognition requires each query patch to recognize all homologous patches within a mixed image.
Due to the \textit{single-centric-object} guarantee~\cite{chen2021multisiam} of ImageNet, that most images are pre-processed to guarantee only one object in the center part of them,
the mixed image can be considered as a ``pseudo'' multi-instance image, 
and given a query patch, the process of recognizing all patches from the same image within a mixed sample is exactly recognizing all patches from the same object within a given ``pseudo'' multi-instance image.
Therefore, the awareness of object existence and completeness is enhanced in the leant representation of our \textit{MixedAE}.

In~\cref{fig:attention_map}, we visualize the attention maps of MAE and \textit{MixedAE} by averaging all attention heads of the last layer, taken the \texttt{[CLS]} token as query and patch tokens as keys.
Compared with MAE which mainly focuses on the most discriminative patches (\eg, boundaries and corners), our \textit{MixedAE} can successfully discover the foreground object patches more precisely and completely, 
which might also explain why \textit{MixedAE} improves more significantly when transferred to dense perception tasks, such as semantic segmentation~\cite{ADE20K} and object detection~\cite{lin2014microsoft}, as shown in~\cref{tab:main_transfer}.


\section{Experiments}

\subsection{Implementation Details}\label{sec:implementation}

\paragraph{Architecture.}

We mainly use the standard ViT-Base~\cite{VIT} as the backbone architecture, 
and further provide ViT-Large experiments in~\cref{app:exp}.
The input images are resized to $224 \times 224$, resulting in a total sequence length $L=196$ with the patch size being $16 \times 16$.
Following MAE~\cite{MAE}, the decoder consists of 8 Transformer layers with the hidden dimension as 512 by default.
For a fair comparison with BEiT~\cite{BEIT}, we additionally build a \textit{MixedAE} in full mixing mode with a lightweight decoder made up of 2 Transformer layers and the hidden dimension as 256, as shown in~\cref{tab:main_transfer}.

The mixing ratio $r$ is set to be 0.25 (\ie, corresponding to the 0.75 masking ratio in MAE~\cite{MAE}) by default, and the threshold $K$ in the $\operatorname{TopK}(\cdot)$ operation is therefore set to be $0.25L$.
Following common practices~\cite{MOCOV3,DINO}, we adopt a linear projector with the output dimension as 128 right before the homologous contrastive loss, and the temperature coefficient $\tau$ is set to be 0.25.


\begin{table*}[t]
\begin{center}
\setlength{\tabcolsep}{0.92mm}
\begin{tabular}{l|cc|c|c|ccc|ccc}
\toprule
\multirow{2}*{Method} & Pre-train & Pre-train$^\dag$ & ImageNet & ADE20K & \multicolumn{6}{c}{COCO} \\
\cline{4-11}
 & Epochs & GPU-days & Top-1 Acc. & mIoU & AP$^{bb}$ & AP$^{bb}_{50}$ & AP$^{bb}_{75}$ & AP$^{mk}$ & AP$^{mk}_{50}$ & AP$^{mk}_{75}$ \\
\hline 
DeiT~\cite{touvron2021training} & 300 & 19.6 & 81.8 & 46.9 & 48.8 & 68.7 & 52.7 & 42.5 & 65.9 & 45.5 \\
MoCov3~\cite{MOCOV3} & 600$^{\dag\dag}$ & 54.8 & 82.8 & 46.8 & 47.2 & 66.9 & 50.8 & 41.1 & 63.6 & 44.1 \\
DINO~\cite{DINO} & 1600$^{\dag\dag}$ & 120.5 & 82.8 & 46.9 & 49.5 & 69.1 & 53.6 & 42.9 & 66.0 & 46.3 \\
\hline 
BEiT~\cite{BEIT} & 300 & 32.1 & 82.9 & 44.7 & 39.3 & 57.7 & 42.4 & 34.8 & 55.2 & 36.8 \\
MAE~\cite{MAE} & 300 & 16.4 & 82.7 & 46.1 & 47.2 & 65.8 & 51.3 & 41.1 & 62.9 & 44.4 \\
MixMIM~\cite{liu2022mixmim} & 300 & 40.2 & 83.2 & - & - & - & - & - & - & -	\\
CIM-RevDet~\cite{fang2022corrupted} & 300 & 42.7 & 83.1 & - & - & - & - & - & - & -	\\	
CIM-ResPix~\cite{fang2022corrupted} & 300 & 42.7 & 83.3 & - & - & - & - & - & - & -	\\
\rowcolor{backcolor}
\textbf{MixedAE} & 300 & 16.9 & 83.1$^{+0.4}$ & 47.0$^{+0.9}$ & 47.8$^{+0.6}$ & 66.6$^{+0.8}$ & 52.0$^{+0.7}$ & 41.6$^{+0.5}$ & 63.6$^{+0.7}$ & 45.0$^{+0.6}$ \\
\rowcolor{backcolor}
\textbf{MixedAE-Full$^*$} & 300 & 30.8 & 83.7$^{+1.0}$ & 47.4$^{+1.3}$ & 48.9$^{+1.7}$ & 67.6$^{+1.8}$ & 53.3$^{+2.0}$ & 42.5$^{+1.4}$ & 64.8$^{+1.9}$ & 45.9$^{+1.5}$ \\
\rowcolor{backcolor}
\textbf{MixedAE-Full} & 300 & 62.3 & \textbf{83.8}$^{+1.1}$ & \textbf{48.9}$^{+2.8}$ & \textbf{51.0}$^{+3.8}$ & \textbf{69.7}$^{+3.9}$ & \textbf{55.2}$^{+3.9}$ & \textbf{44.1}$^{+3.0}$ & \textbf{67.0}$^{+4.1}$ & \textbf{47.9}$^{+3.5}$ \\
\hline 
BEiT~\cite{BEIT} & 800 & 85.5 & 83.2 & 45.6 & 40.8 & 59.4 & 44.1 & 36.0 & 56.8 & 38.2 \\
MAE~\cite{MAE} & 800 & 43.7 & 83.3 & 47.2 & 49.4 & 68.1 & 53.9 & 42.9 & 65.5 & 46.6 \\
\rowcolor{backcolor}
\textbf{MixedAE} & 800 & 45.0 & \textbf{83.5}$^{+0.2}$ & \textbf{48.7}$^{+1.5}$ & \textbf{50.3}$^{+0.9}$ & \textbf{69.1}$^{+1.0}$ & \textbf{54.8}$^{+0.9}$ & \textbf{43.5}$^{+0.6}$ & \textbf{66.2}$^{+0.7}$ & \textbf{47.4}$^{+0.8}$ \\
\hline 
MAE~\cite{MAE} & 1600 & 87.4 & 83.6 & 48.1 & 50.6 & 69.4 & 55.0 & 43.8 & 66.6 & 47.5 \\
iBOT~\cite{ibot} & 1600$^{\dag\dag}$ & 172.1 & 83.8 & 49.6 & 51.2 & 70.1 & 55.2 & 44.3 & 67.4 & 48.0 \\
\rowcolor{backcolor}
\textbf{MixedAE} & 1600 & 90.1 & \textbf{83.9}$^{+0.3}$ & \textbf{49.8}$^{+1.7}$ & \textbf{51.5}$^{+0.9}$ & \textbf{70.2}$^{+0.8}$ & \textbf{55.9}$^{+0.9}$ & \textbf{44.5}$^{+0.7}$ & \textbf{67.5}$^{+0.9}$ & \textbf{48.2}$^{+0.7}$ \\
\bottomrule
\end{tabular}
\end{center}
\vspace{-5mm}
\caption{{\bf Transfer performance comparison between methods pre-trained on ImageNet-1K.}
1) Effectiveness: \textit{MixedAE} achieves the state-of-the-art performance under different pre-training epochs and overheads.
2) Efficiency: \textit{MixedAE} consistently surpasses the strong iBOT~\cite{ibot} baseline, while only requiring 53.4\% of the pre-training overhead.
3) Object-aware pre-training: more significant improvements are achieved when transferred to downstream dense perception tasks (0.3 vs. 1.7 vs. 0.9).
$^*$: a lightweight decoder is deployed to maintain similar pre-training overhead with BEiT~\cite{BEIT}. 
$^\dag$: GPU-days estimated on Tesla V100 GPUs.
$^{\dag\dag}$: effective epochs following iBOT~\cite{ibot}.
}
\vspace{-4mm}
\label{tab:main_transfer}
\end{table*}


\vspace{-3mm}
\paragraph{Optimization.}
We pre-train \textit{MixedAE} on the ImageNet-1K~\cite{deng2009imagenet} training set with the AdamW~\cite{adamw} optimizer and a cosine learning rate schedule with a linear warm-up of 40 epochs. 
The batch size is set to be 4096 for the compose mixing, while 1024 for the full mode.
The base learning rate is set to be $7.5e^{-5}$, which will scale linearly with the batch size ($lr = lr_{base} \times bs / 256$).
Only standard random cropping and flipping are utilized for data augmentation.
The remaining hyperparameters are all maintained the same with MAE for a fair comparison 
(see~\cref{app:implementation} for more details).


\subsection{Transfer Results on ImageNet-1K}\label{sec:transfer_IN}

\paragraph{Setup.}
We consider the fully fine-tuning performance on ImageNet-1K for 100 epochs and report the Top-1 accuracy.
Following MAE~\cite{MAE}, we average all the patch tokens after the final Transformer layer, which is taken as input to a linear head for classification.
See more details in~\cref{app:implementation}.

\vspace{-3mm}
\paragraph{Comparison with MAE.}
As shown in~\cref{tab:main_transfer}, \textit{MixedAE} obtains consistent improvements over MAE under different pre-training epochs with only 3\% additional overhead.
The 300-epoch pre-trained \textit{MixedAE} with full mixing acquires even better accuracy than the 1600-epoch pre-trained MAE, demonstrating the efficiency of data utilization.

\vspace{-3mm}
\paragraph{Comparison with other MIM augmentations.}
Our \textit{MixedAE} with the lightweight decoder and the full mixing mode obtains 83.7\% Top-1 accuracy, 0.5\% and 0.4\% higher than MixMIM~\cite{liu2022mixmim} and CIM~\cite{fang2022corrupted} respectively, meanwhile saving at least 23.4\% computational overhead, revealing the simplicity of our \textit{MixedAE}.

\vspace{-3mm}
\paragraph{Comparison with other SSL approaches.}
Our \textit{MixedAE} obtains consistent improvements under various pre-training epochs and overheads, and achieves a better trade-off between pre-training overhead and transfer performance, as shown in~\cref{fig:tradeoff}.
Specifically, \textit{MixedAE} pre-trained for 1600 epochs achieves 83.9\% accuracy, constructing a new state-of-the-art result with a pure \textbf{autoencoder}-based framework.
Requiring only 53.4\% of the training overhead, \textit{MixedAE} surpasses the strong iBOT~\cite{ibot}, demonstrating remarkable efficiency. 
Moreover, the improvement brought by mixing is orthogonal to the usage of more advanced reconstruction targets~\cite{PeCo,MaskFeat,data2vec} and masking strategies~\cite{kakogeorgiou2022hide,shi2022adversarial}.


\subsection{Transfer Results on Downstream Tasks}\label{sec:transfer_downstream}

We further consider three downstream settings to evaluate the learnt representation, 
and more details about the different transfer procedures are included in~\cref{app:implementation}.

\vspace{-3mm}
\paragraph{Semantic segmentation.}
We utilize the UperNet~\cite{xiao2018unified} to perform semantic segmentation on ADE20K~\cite{ADE20K} following BEiT~\cite{BEIT}.
As reported in~\cref{tab:main_transfer}, our 800-epoch \textit{MixedAE} achieves 48.7 mIoU, even surpassing the MAE pre-trained for 1600 epochs by 0.6 mIoU, and our 1600-epoch \textit{MixedAE} further improves to 49.8 mIoU, outperforming all baseline methods by a non-trivial margin, which is more significant than the improvement on ImageNet-1K (1.7 vs. 0.3), thanks to the object-aware pre-training, as discussed in~\cref{sec:discussion}.

\vspace{-3mm}
\paragraph{Object detection and instance segmentation.}
We utilize Cascade Mask R-CNN~\cite{cai18cascadercnn,he2017mask} to produce bounding boxes and instance masks simultaneously on COCO~\cite{lin2014microsoft}.
As demonstrated in~\cref{tab:main_transfer}, \textit{MixedAE} consistently outperforms MAE under different epochs (0.6/0.9/0.9 \& 0.5/0.6/0.7).
Similarly with ADE20K, more significant improvements are observed due to the high-quality attention maps learnt by the object-aware pre-training, as demonstrated in~\cref{fig:attention_map}.


\begin{table*}[t!]
    \centering
    \begin{center}
    \scalebox{1.0}{
        \setlength{\tabcolsep}{1.8mm}
        \begin{tabular}{lcccccccccccc}
        \toprule
        Method & Aircraft & Caltech & Cars & C10 & C100 & DTD & Flowers & Food & Pets & SUN & VOC & Avg. \\
        \midrule
        \multicolumn{13}{l}{\textit{SSL ResNets}} \\
        MoCov2~\cite{chen2020improved} & 79.9 & 84.4 & 75.2 & 96.5 & 71.3 & 69.5 & 94.4 & 76.8 & 79.8 & 55.8 & 71.7 & 77.7 \\
        SimCLR~\cite{chen2020simple} & 78.7 & 82.9 & 79.8 & 96.2 & 79.1 & 70.2 & 94.3 & 82.2 & 83.2 & 61.1 & 78.2 & 80.5 \\
        BYOL~\cite{grill2020bootstrap} & 79.5 & 89.4 & 84.6 & 97.0 & 84.0 & 73.6 & 94.5 & 85.5 & 89.6 & 64.0 & 82.7 & 84.0 \\
        SwAV~\cite{caron2020unsupervised} & \textbf{83.1} & 89.9 & 86.8 & 96.8 & 84.4 & 75.2 & 95.5 & 87.2 & 89.1 & 66.2 & 84.7 & 85.3 \\
        SDR~\cite{liu2022task} & 82.6  & 89.0  & 87.5  & 97.4  & 84.4  & 75.6  & 97.0  & 86.1  & 89.3  & 66.1  & 85.3  & 85.5  \\
        \hline
        \multicolumn{13}{l}{\textit{SSL Transformers}} \\
        MoCov3~\cite{MOCOV3} & 76.6 & 91.2 & 86.6 & 98.3 & 88.3 & 72.6 & 95.5 & 86.4 & 92.0 & 65.6 & 84.5 & 85.2 \\
        DINO~\cite{DINO} & 69.4 & 91.2 & 81.3 & \textbf{98.4} & \textbf{88.9} & 77.6 & 96.9 & 87.3 & 93.5 & 64.7 & 86.3 & 85.1 \\
        BEiT~\cite{BEIT} & 66.3 & 80.2  & 78.6  & 96.1 & 80.0 & 69.9  & 92.9 & 83.2 & 85.3 & 57.1 & 76.7 & 78.7  \\
        MAE~\cite{MAE} & 78.2 & 91.2 & 88.4 & 97.0 & 82.5 & 75.3 & 96.6 & 84.7 & 92.6 & 65.4 & 86.0 & 85.3 \\
        \hline
        \rowcolor{backcolor}
        \textbf{MixedAE} & 82.1 & \textbf{91.5} & \textbf{88.8} & 97.9 & 85.9 & \textbf{78.7} & \textbf{97.1} & \textbf{87.4} & \textbf{93.6} & \textbf{66.2} & \textbf{86.4} & \textbf{86.9}$^{+1.6}$ \\
        \bottomrule
        \end{tabular}
    }
\end{center}
\vspace{-4mm}
\caption{\textbf{Transfer performance comparison on 11 downstream classification tasks}. 
Our 1600-epoch pre-trained \textit{MixedAE} achieves consistent improvements over MAE on all 11 classification datasets with an average accuracy of 86.9\%, surpassing all counterparts.}
\vspace{-4mm}
\label{tab:fine_transfer}
\end{table*}


\paragraph{Downstream classification.}
Following~\cite{ericsson2021well,liu2022task,zhili2023task}, we study transfer performance on 11 downstream classification datasets, including both fine-grained (\eg, Cars~\cite{krausecollecting}) and coarse-grained ones (\eg, CIFAR100~\cite{krizhevsky2009learning}) in~\cref{tab:fine_transfer}.
Our \textit{MixedAE} achieves consistent improvement over MAE on all 11 downstream tasks with an average accuracy of 86.9\%, outperforming all counterparts as demonstrated in~\cref{tab:fine_transfer}.


\begin{table*}[t]
\vspace{-.2em}
\centering
\subfloat[
    \textbf{Mixing ratio $r$}, which is not limited to 0.5, different from the MixMIM~\cite{liu2022mixmim} formulation.
    \label{tab:mixing_ratio}
    ]{
    \centering
    \begin{minipage}{0.29\linewidth}{
        \begin{center}
        \tablestyle{1pt}{1.05}
        \begin{tabular}{x{48}x{40}x{24}x{24}}
            mixing ratio $r$ & GPU-days & acc & mIoU \\
            \shline
            0.25 & \baseline{\textbf{16.9}} & \baseline{\textbf{82.4}} & \baseline{\textbf{45.0}} \\
            0.5 & 21.0 & 82.3 & 44.2 \\
        \end{tabular}
        \end{center}
    }
    \end{minipage}
}
\hspace{2em}
\subfloat[
    \textbf{Position of homologous contrastive}.
    The encoder features after \texttt{[LN]} works better.
    \label{tab:homo_position}
    ]{
    \begin{minipage}{0.29\linewidth}{
        \begin{center}
        \tablestyle{4pt}{1.05}
        \begin{tabular}{y{56}x{24}x{24}}
            case & acc & mIoU \\
            \shline
            pre-\texttt{[LN]} & 82.4 & 44.5 \\
            post-\texttt{[LN]} & \baseline{\textbf{82.5}} & \baseline{\textbf{45.3}} \\
        \end{tabular}
        \end{center}
    }
    \end{minipage}
}
\hspace{2em}
\subfloat[
    \textbf{Positives of homologous contrastive}.
    It's better to utilize positive patches separately.
    \label{tab:homo_positive}
    ]{
    \begin{minipage}{0.29\linewidth}{
        \begin{center}
        \tablestyle{4pt}{1.05}
        \begin{tabular}{y{75}x{24}x{24}}
            case & acc & mIoU \\
            \shline
            all positives separately & \baseline{\textbf{82.5}} & \baseline{\textbf{45.3}} \\
            mean positive & 82.5 & 45.1 \\
        \end{tabular}
        \end{center}
        }
    \end{minipage}
} \\
\centering
\vspace{.3em}
\subfloat[
    \textbf{Threshold $K$} works best when consistent with the mixing ratio $r$ in~\cref{tab:mixing_ratio}.
    \label{tab:attn_K}
    ]{
    \begin{minipage}{0.29\linewidth}{
        \begin{center}
        \tablestyle{4pt}{1.05}
        \begin{tabular}{x{64}x{24}x{24}}
            threshold $K$ & acc & mIoU \\
            \shline
            0.125$L$ & 82.6 & 45.1 \\
            0.25$L$ & \baseline{\textbf{82.6}} & \baseline{\textbf{45.6}} \\
            0.5$L$ & 82.6 & 45.5\\
            1.0$L$ (no sampling) & 82.5 & 45.3 \\
        \end{tabular}
        \end{center}
    }
    \end{minipage}
}
\hspace{2em}
\subfloat[
    \textbf{Usage of global self-attention} in the very first layer only achieves the best performance.
    \label{tab:attn_layer}
    ]{
    \centering
    \begin{minipage}{0.29\linewidth}{
        \begin{center}
        \tablestyle{4pt}{1.05}
        \begin{tabular}{x{56}x{24}x{24}}
            layer ID & acc & mIoU \\
            \shline
            none & 82.6 & 45.6 \\
            1st & \baseline{\textbf{82.7}} & \baseline{\textbf{46.4}} \\
            1st \& 2nd & 82.5 & 44.8 \\
            1st \& 2nd \& 3rd & 82.5 & 45.6 \\
        \end{tabular}
        \end{center}
        }
    \end{minipage}
}
\hspace{2em}
\subfloat[
    \textbf{Homologous recognition}.
    Best to utilize homologous attention and contrastive together.
    \label{tab:homo_summary}
    ]{
    \begin{minipage}{0.29\linewidth}{
        \begin{center}
        \tablestyle{1pt}{1.05}
        \begin{tabular}{y{24}x{32}x{32}x{24}x{24}}
        & HomoAtt & HomoCon & acc & mIoU \\
        \shline
        \multirow{4}*{\textit{MixedAE}} & & & 82.4 & 45.0 \\
         & \checkmark & & 82.6 & 44.5 \\
         & & \checkmark & 82.5 & 45.3 \\
         & \checkmark & \checkmark & \baseline{\textbf{82.7}} & \baseline{\textbf{46.4}} \\
        \end{tabular}
        \end{center}
    }\end{minipage}
}
\vspace{-1.em}
\caption{\textbf{\textit{MixedAE} ablation experiments} with ViT-B/16. 
We report the ImageNet-1K fine-tuning accuracy (acc) and ADE20K semantic segmentation performance (mIoU).
We explore (a) the mixing ratio $r$, (b,c) the position and positives for homologous contrastive, (d,e) the threshold $K$ and position for homologous attention, and (f) a summary of homologous recognition.
Default settings are marked in \colorbox{baselinecolor}{gray}.}
\label{tab:ablation}  
\vspace{-.5em}
\end{table*}


\begin{figure*}[!thbp]
  \centering
  \begin{subfigure}{0.49\linewidth}
    \includegraphics[width=1.0\linewidth]{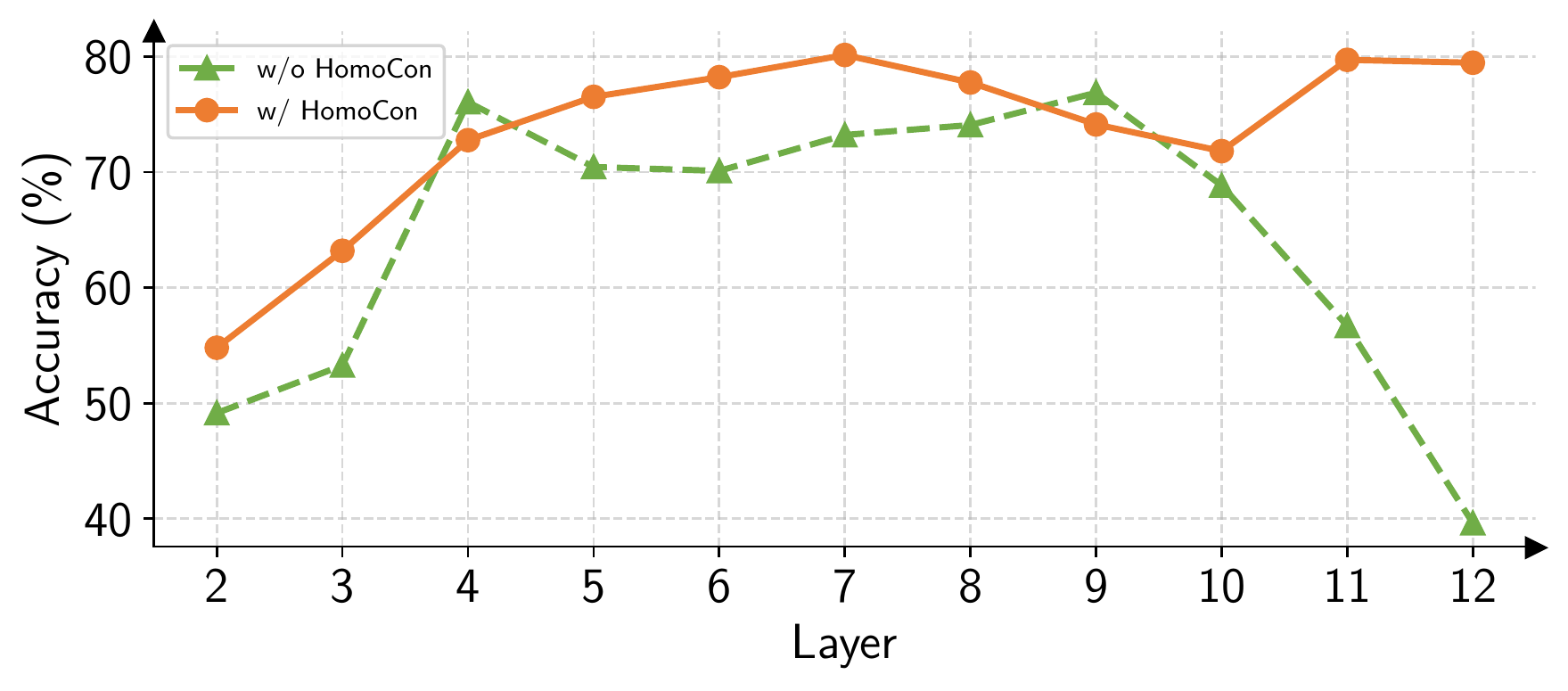}
    \caption{\textbf{Accuracy curve with respect to Transformer layers} of our \textit{MixedAE} pre-trained for 300 epochs, where the 1st layer adopts global self-attention.}
    \label{fig:layer}
  \end{subfigure}
  \hfill
  \begin{subfigure}{0.49\linewidth}
    \includegraphics[width=1.0\linewidth]{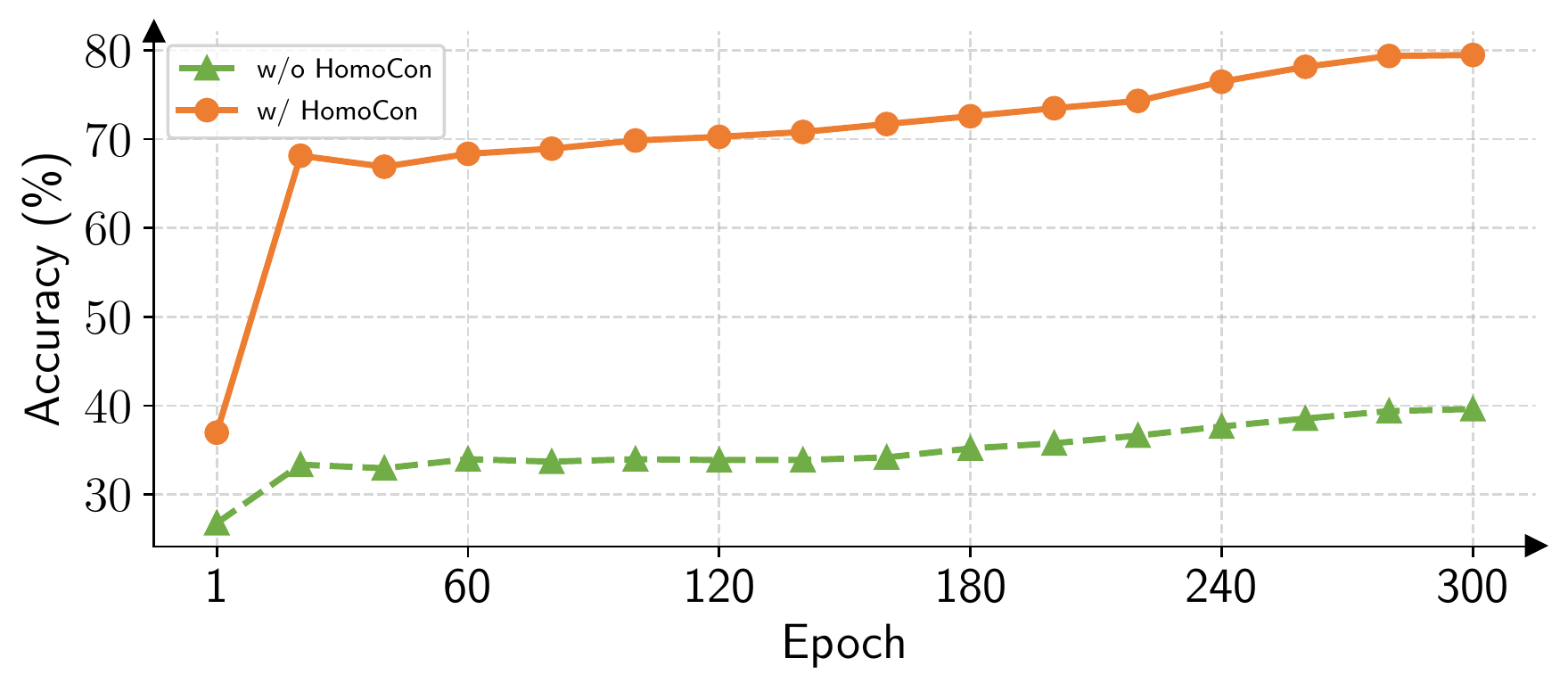}
     \caption{\textbf{Accuracy curve with respect to pre-training epochs} of the very final Transformer layer in our \textit{MixedAE}.}
    \label{fig:epoch}
  \end{subfigure}
  \vspace{-2mm}
  \caption{\textbf{Analysis of $\operatorname{TopK}$ sampling accuracy}.
  Without the homologous contrastive loss for verification (\textcolor{G}{green} curve), utilizing the homologous attention only cannot achieve promising accuracy, dramatically varying between different layers.
  However, with the usage of homologous contrastive loss (\textcolor{O}{orange} curve), the sampling accuracy is significantly improved and stabilized mostly around 70\% to 80\% throughout the whole pre-training process, which is essential to achieve remarkable transfer performance, as demonstrated in~\cref{tab:homo_summary}.
  }
  \vspace{-3mm}
  \label{fig:accuracy}
\end{figure*}



\subsection{Ablation Study}\label{sec:ablation}

\paragraph{Setup.}
We conduct 300-epoch pre-training with a base learning rate of $1.5e^{-4}$ for all ablation studies on \textit{MixedAE} with compose mixing.
By default, we report the fine-tuning accuracy on ImageNet-1K~\cite{deng2009imagenet} and mIoU on ADE20K~\cite{ADE20K}.
See more detailed settings and results in~\cref{app:exp}.

\vspace{-3mm}
\paragraph{Mixing ratio $r$.}
Different from MixMIM~\cite{liu2022mixmim}, the mixing ratio $r$ in our formulation can be flexibly selected from the range of $(0,0.5]$.
As shown in~\cref{tab:mixing_ratio}, $r=0.25$ works better, while requiring less pre-training overhead (since the effective encoder batch size scales linearly with $r$, as shown in~\cref{equ:mixing}).
Note that $r=0.25$ is also corresponding to the default 0.75 masking ratio in MAE.

\vspace{-3mm}
\paragraph{Position of homologous contrastive.}
We study whether the encoder features before or after the final Layer Normalization (\texttt{[LN]})~\cite{LayerNorm} of ViT~\cite{VIT} achieves better performance as input to the homologous contrastive loss in~\cref{tab:homo_position}.
The latter achieves consistent improvements on both ImageNet-1K and ADE20K, suggesting that the features after \texttt{[LN]} are more suitable for homologous recognition.

\vspace{-3mm}
\paragraph{Positives of homologous contrastive.}
In~\cref{equ:homo_con}, given a query patch, all homologous patches are considered as positive samples but taken separately to calculate $\mathcal{L}_{HomoCon}$ in the supervised contrastive manner~\cite{khosla2020supervised}.
We further ablate to utilize the average of both the query and all its homologous patches as its positive, which, however, performs slightly worse than the separate manner (-0.2 mIoU on ADE20K).

\paragraph{Threshold $K$ of homologous attention.}
We study the threshold number $K$ in $A_{HomoAtt}$ in~\cref{tab:attn_K}, where \textbf{all} the global self-attention operations in the ViT are replaced with our homologous attention.
Compared with the no sampling baseline, the usage of homologous attention obtains consistent improvement on ImageNet-1K, while the best achieves at $K=0.25L$, which is consistent with the mixing ratio $r$ in~\cref{tab:mixing_ratio} specifically.

\vspace{-3mm}
\paragraph{Position of homologous attention.}
As shown in~\cref{fig:layer}, homologous attention cannot achieve promising accuracy in early Transformer layers without sufficient information engagement.
Thus, we further explore to maintain the global self-attention in early layers unchanged in~\cref{tab:attn_layer}, and observe empirically that utilizing global self-attention in the very first layer only achieves the best performance.

\vspace{-3mm}
\paragraph{Homologous recognition.}
In~\cref{tab:homo_summary}, we further compare the effectiveness of different components of homologous recognition.
Without the homologous contrastive loss for verification, utilizing homologous attention only obtains a significant drop of 0.5 mIoU on ADE20K.
Although achieving improvement, utilizing the homologous contrastive only still suffers from the ease of reconstruction brought by MI increasement, as previously discussed in~\cref{equ:mi}.
Finally, the best performance is achieved when using homologous attention and contrastive loss simultaneously.


\subsection{Analysis}\label{sec:analysis}

\paragraph{Effectiveness of $\operatorname{TopK}$ sampling.}
To further observe the effectiveness of $\operatorname{TopK}$ sampling in homologous attention, we visualize the sampling accuracy with respect to different layers and pre-training epochs in~\cref{fig:accuracy}.
As demonstrated in~\cref{fig:layer}, the na\"ive usage of homologous attention only cannot achieve promising sampling accuracy, which, therefore, suffers from a significant performance drop in~\cref{tab:homo_summary}.
Specifically, neither the sampling accuracy of the first two nor final two layers exceed 60\%, and accuracy of the very final layer even maintains under 40\% throughout the whole pre-training process, as shown in~\cref{fig:epoch} (the \textcolor{G}{green} curve).

However, with the usage of the homologous contrastive loss for verification, the sampling accuracy is significantly improved and stabilized around 70\% to 80\% for all layers except the first two, as in~\cref{fig:layer}.
The sampling accuracy of the final layer rapidly increases to around 70\% when pre-trained only for 20 epochs, maintaining stable throughout the remaining pre-training, as in~\cref{fig:epoch} (the \textcolor{O}{orange} curve).

\vspace{-3mm}
\paragraph{Comparison with existing MIM methods combined with contrastive.} 
Although also utilizing a ``contrastive loss'' with reconstruction, \textit{MixedAE} differs from existing MIM works~\cite{el2021large,ibot} combined with contrastive learning from two perspectives,
1) \textbf{Purpose}: existing works utilize contrastive loss to perform instance discrimination simultaneously with MIM, while our homologous contrastive is only utilized to guarantee the sampling accuracy.
Therefore, homologous contrastive performs more like a regularization term instead of an individual self-supervision in~\cite{el2021large,ibot}.
To verify, we pre-train a \textit{MixedAE} with $\mathcal{L}_{HomoCon}$ only without $\mathcal{L}_{recon}$, which cannot achieve reasonable performance, 
as reported in~\cref{app:exp}.
2) \textbf{Efficiency}: given a single input, existing works require forward propagation at least twice for several augmented views to conduct instance discrimination, while only once for our homologous contrastive, resulting in the significant efficiency.
Specifically, our \textit{MixedAE} surpasses iBOT~\cite{ibot} with only 53.4\% of its computational overhead.


\section{Conclusion}
This paper explores the usage of image mixing for MAE.
Different from in supervised and contrastive learning, we first theoretically demonstrate na\"ive mixing might instead ease the reconstruction pretext task. 
To address that, our \textit{MixedAE} with the proposed homologous recognition as the auxiliary supervision can not only achieve state-of-the-art performance with a better trade-off between transfer results and pre-training overhead, but also conduct object-aware pre-training without any specifically designed modules.
We hope our simple yet effective method can bring researchers' attention to more effective data augmentations for MIM.

\vspace{-3mm}
\paragraph{Acknowledgments.}
We gratefully acknowledge support of MindSpore, CANN (Compute Architecture for Neural Networks) and Ascend AI Processor used for this research.

{\small
\bibliographystyle{ieee_fullname}
\bibliography{egbib}
}


\clearpage

\onecolumn
\appendix

\section*{Appendix}

\section{Proof of~\cref{equ:mi}}\label{app:mi_proof}
In this section, we prove~\cref{equ:mi} with the terminologies maintained consistent with~\cref{sec:method}. 
We start to prove when $r=0.5$ (\ie, two clean images within a single group).
Denote $\boldsymbol{X}_1,\boldsymbol{X}_2$ as two random variables representing two input images, and $\boldsymbol{M}$ as the random mask, which can be considered as a constant here since it is independently generated with $\boldsymbol{X}_1$ and $\boldsymbol{X}_2$.
Then, according to~\cref{equ:mixing}, the mixed input can be represented as,
\begin{equation}
    \sigma_{mix}(\{\boldsymbol{X}_1, \boldsymbol{X}_2\}, \boldsymbol{M}) = \mathbbm{1}(\boldsymbol{M} = 1)\boldsymbol{X}_1 + \mathbbm{1}(\boldsymbol{M} = 2)\boldsymbol{X}_2,
    \label{equ:mix}
\end{equation}
while the MAE input can be represented as,
\begin{equation}
    \sigma_{MAE}(\boldsymbol{X}_1, \boldsymbol{M}) = \{\boldsymbol{X}_{1,l}|M_l=1\} = \mathbbm{1}(\boldsymbol{M} = 1)\boldsymbol{X}_1 + \mathbbm{1}(\boldsymbol{M} = 2)\boldsymbol{\overrightarrow{0}}.
\end{equation}
Therefore, given $\boldsymbol{X}_1$ as the reconstruction target, we can represent the mutual information (MI) between the mixed input and the reconstruction target $\boldsymbol{X}_1$ as,
\begin{equation}
\begin{aligned}
    I(\sigma_{mix}(\{\boldsymbol{X}_1, \boldsymbol{X}_2\}, \boldsymbol{M}); \boldsymbol{X}_1) & = I(\mathbbm{1}(\boldsymbol{M} = 1)\boldsymbol{X}_1 + \mathbbm{1}(\boldsymbol{M} = 2)\boldsymbol{X}_2;\boldsymbol{X}_1)\\
    & = H(\boldsymbol{X}_1) - H(\boldsymbol{X}_1|\mathbbm{1}(\boldsymbol{M} = 1)\boldsymbol{X}_1 + \mathbbm{1}(\boldsymbol{M} = 2)\boldsymbol{X}_2) \\
    & = H(\boldsymbol{X}_1) - H(\boldsymbol{X}_1|\mathbbm{1}(\boldsymbol{M} = 1)\boldsymbol{X}_1 + \mathbbm{1}(\boldsymbol{M} = 2)\boldsymbol{\overrightarrow{0}}+ \mathbbm{1}(\boldsymbol{M} = 1)\boldsymbol{\overrightarrow{0}} + \mathbbm{1}(\boldsymbol{M} = 2)\boldsymbol{X}_2) \\
    & = H(\boldsymbol{X}_1) - H(\boldsymbol{X}_1|\mathbbm{1}(\boldsymbol{M} = 1)\boldsymbol{X}_1 + \mathbbm{1}(\boldsymbol{M} = 2)\boldsymbol{\overrightarrow{0}},  \mathbbm{1}(\boldsymbol{M} = 1)\boldsymbol{\overrightarrow{0}} + \mathbbm{1}(\boldsymbol{M} = 2)\boldsymbol{X}_2) \\
    & \ge H(\boldsymbol{X}_1) - H(\boldsymbol{X}_1|\mathbbm{1}(\boldsymbol{M} = 1)\boldsymbol{X}_1 + \mathbbm{1}(\boldsymbol{M} = 2)\boldsymbol{\overrightarrow{0}}) \\
    & = I(\sigma_{MAE}(\boldsymbol{X}_1 ,\boldsymbol{M});\boldsymbol{X}_1),
    \label{equ:mi_conclusion}
\end{aligned}
\end{equation}
where $H(\cdot)$ is the entropy.
The conclusion above also holds when $\boldsymbol{X}_2$ is considered as the reconstruction target symmetrically.

Note that although independent in the data space, $\boldsymbol{X}_1$ and $\boldsymbol{X}_2$ are not independent in the feature space, because the \textbf{global self-attention} would introduce inevitable information leakage from $\boldsymbol{X}_2$ to $\boldsymbol{X}_1$, which would be enhanced for ``relevant'' $\boldsymbol{X}_2$ 
(\eg, the green cucumber in~\cref{fig:uncertainty}),
while restrained for ``irrelevant'' ones (\eg, the blue sky) by assigning different attention weights. 
As shown in~\cref{fig:accuracy}, 
the $\operatorname{TopK}(\cdot)$ sampling accuracy converges to around 80\%, suggesting that the information leakage does exist in practice.
Therefore, the independence is affected, and the equality in~\cref{equ:mi_conclusion} does not hold.

For $r\in (0, 0.5)$, there are more than two images within a single group $\{\boldsymbol{X}_1, \boldsymbol{X}_2,...,\boldsymbol{X}_{1/r}\}$.
Considering $\boldsymbol{X}_1$ as the reconstruction target, we can first mix all images except $\boldsymbol{X}_1$ to generate a pseudo $\boldsymbol{\hat{X}}_2$ as,
\begin{equation}
    \boldsymbol{\hat{X}}_2 = \sigma_{mix}(\{\boldsymbol{\overrightarrow{0}},\boldsymbol{X}_2,\boldsymbol{X}_3,...,\boldsymbol{X}_{1/r}\}, \boldsymbol{M}) = \mathbbm{1}(\boldsymbol{M} = 1)\boldsymbol{\overrightarrow{0}} + \sum_{i=2}^{1/r}\mathbbm{1}(\boldsymbol{M} = i)\boldsymbol{X}_i,
\end{equation}
which is then mixed with $\boldsymbol{X}_1$ following~\cref{equ:mix}.
Therefore, the conclusion in~\cref{equ:mi_conclusion} can still be satisfied.

As discussed above, the usage of the \textbf{global self-attention} in ViT~\cite{VIT} is another indispensable factor to achieve the MI increasement in~\cref{equ:mi_conclusion}.
Therefore, in this paper, we propose the homologous attention to replace the global self-attention together with the homologous contrastive loss as verification for our \textit{MixedAE}.


\section{More Implementation Details}\label{app:implementation}


\begin{table}[!tb]
    \begin{minipage}{.49\linewidth}
        \centering
        \begin{tabular}{l|l}
            \toprule
            config & value \\
            \hline
            optimizer & AdamW~\cite{adamw} \\
            base learning rate & $7.5e^{-5}$ \\
            weight decay & 0.05 \\
            optimizer momentum & $\beta_1,\beta_2=0.9,0.95$  \\
            batch size & 4096 (B), 2048 (L) \\
            learning rate schedule & cosine decay~\cite{Loshchilov2016} \\
            warmup epochs & 40 \\
            augmentation & RandomResizedCrop \\
            reconstruction target & normalized pixels~\cite{MAE} \\
            \bottomrule
            \multicolumn{2}{c}{~}\\
            \multicolumn{2}{c}{~}\\
            \multicolumn{2}{c}{~}\\
            \multicolumn{2}{c}{~}\\
        \end{tabular}
        \caption{\textbf{Pre-training settings.}}
        \label{tab:param_pretrain}
    \end{minipage}
    \begin{minipage}{.49\linewidth}
        \centering
        \begin{tabular}{l|l}
            \toprule
            config & value \\
            \hline
            optimizer & AdamW \\
            base learning rate & $5e^{-4}$ (B), $1e^{-3}$ (L) \\
            weight decay & 0.05 \\
            optimizer momentum & $\beta_1, \beta_2=0.9, 0.999$ \\
            batch size & 1024 (B), 512 (L) \\
            learning rate schedule & cosine decay \\
            warmup epochs & 5 \\
            training epochs & 100 (B), 50 (L) \\
            augmentation & RandAug (9, 0.5) \\
            label smoothing \cite{Szegedy2016a} & 0.1 \\
            mixup \cite{Mixup} & 0.8 \\
            cutmix \cite{CutMix} & 1.0 \\
            drop path \cite{Huang2016} & 0.1 \\
            \bottomrule
        \end{tabular}
        \caption{\textbf{Fully fine-tuning settings.}}
        \label{tab:param_IN}
    \end{minipage}
\end{table}


\paragraph{Pre-training.}
The default settings are provided in~\cref{tab:param_pretrain}.
We use xavier\_uniform~\cite{Glorot2010} to initialize all Transformer layers, following the official code of ViT~\cite{VIT}.
Normalized pixels~\cite{MAE} are utilized as the reconstruction target, and the mask patch size is set to be $32\times 32$, following~\cite{SimMIM,liu2022mixmim}.
In practice, we utilize the sine-cosine encodings~\cite{vaswani2017attention} for segment embeddings, which is added to the input of each Transformer layer following~\cite{liu2021swin}.
If no otherwise specified, the compose mixing mode is by default adopted to generate a single mixed sample for each image group.

\paragraph{ImageNet classification.}
The default settings are provided in~\cref{tab:param_IN}.
We mainly adopt the fully fine-tuning transfer setting to fine-tune the parameters of the backbone and the classification head simultaneously.
We utilize the layer-wise learning rate decay strategy~\cite{electra} following~\cite{BEIT}. 
In practice, we sweep the decay ratio in $\{0.65, 0.7, 0.75\}$ following~\cite{ibot,liu2022mixmim}.

\paragraph{ADE20K semantic segmentation.}
We use UperNet~\cite{xiao2018unified} as the segmentor following BEiT~\cite{BEIT}.
The input resolution is $512\times 512$, and the batch size is set to be 16.
The learning rate is set to be $3e^{-4}$ with the layer-wise learning rate decay ratio as 0.65 for ViT-Base and 0.75 for ViT-Large.
We conduct fine-tuning for 160K iterations, and evaluate the performance without the multi-scale augmentation.

\paragraph{COCO object detection and instance segmentation.}
We utilize the Cascade Mask R-CNN~\cite{cai18cascadercnn,he2017mask} following iBOT~\cite{ibot}.
Multi-scale training is adopted with the shorted side randomly resized between 480 and 800 while the longer side no larger than 1333.
The batch size is 16, the initial learning rate is $1e^{-4}$, and the layer-wise learning rate decay ratio is set to be 0.75.
We adopt the standard $1\times$ schedule to train for 12 epochs, and decrease the learning rate by $10\times$ at epoch 9 and 11.

\paragraph{Downstream classification.}
We mainly follow the setups in~\cite{ liu2022task,zhili2023task} to evaluate the transfer performance on 11 downstream classification datasets, 
including both the fine-grained datasets (\eg, Aircraft~\cite{maji2013fine}, Cars~\cite{krausecollecting}, Flowers~\cite{nilsback2008automated}, Food~\cite{bossard14}, Pets~\cite{parkhi2012cats} and SUN397~\cite{xiao2010sun}), 
and the coarse-grained ones (\eg, Caltech101~\cite{fei2004learning}, CIFAR10~\cite{krizhevsky2009learning}, CIFAR100~\cite{krizhevsky2009learning}, DTD~\cite{cimpoi2014describing} and VOC2007~\cite{everingham2010pascal}). 
Specifically, we adopt a linear classification head upon the pre-trained ViT backbone and fully fine-tune the whole model simultaneously for 5000 iterations. 
The SGD optimizer and the cosine learning rate schedule are adopted. 
We grid search the optimal learning rate among $\{1e^{-3}, 3e^{-3}, 1e^{-2}, 3e^{-2}\}$, and set the weight decay to be 0. 


\section{More Experiments}\label{app:exp}

\paragraph{Scaling property of \textit{MixedAE}.}
We further build \textit{MixedAE} with the standard ViT-Large~\cite{VIT} as the backbone architecture, and pre-train for 1600 epochs on ImageNet-1K~\cite{deng2009imagenet} following the same optimization recipe with ViT-Base as in~\cref{app:implementation}.
As demonstrated in~\cref{tab:vit_large}, \textit{MixedAE} still outperforms MAE consistently with ViT-Large, especially on downstream dense perception tasks~\cite{ADE20K,lin2014microsoft,han2021soda10m,li2022coda} thanks to the object-aware pre-training, revealing the scalability of our proposed \textit{MixedAE}.

Moreover, we further report the performance of the 1500-epoch pre-trained \textit{MixedAE} with ViT-Large to maintain similar computational overhead with MAE in~\cref{tab:vit_large}.
\textit{MixedAE} still obtains consistent improvements over MAE, while outperforming iBOT with a 1.8$\times$ acceleration, suggesting that \textit{MixedAE} can achieve a better trade-off between the computational overhead and the transfer performance regardless of the architecture size.

\paragraph{Ablation settings in~\cref{tab:ablation}.} 
We conduct the ablation of new components based on previous results.
Starting from the na\"ive baseline in~\cref{sec:baseline}, we first ablate the mixing ratio $r$ 
in~\cref{tab:mixing_ratio}.
Accordingly, we fix $r = 0.25$ to explore the position and positives of the homologous contrastive in~\cref{tab:homo_position,tab:homo_positive}.
Similarly, we ablate homologous attention in~\cref{tab:attn_K,tab:attn_layer} based on results of~\cref{tab:homo_positive}, and finally summarize all components in~\cref{tab:homo_summary}.

\paragraph{Main cause of performance degradation of na\"ive mixing.}
To verify that the information leakage brought by the mutual information increasement, as proved in~\cref{app:mi_proof}, is indeed the main cause of performance degradation of the na\"ive mixing baseline introduced 
in~\cref{sec:baseline}
instead of the optimization difficulty, we further build a mixing baseline by:
1) applying the masked self-attention to perform self-attention only within homologous patches to prevent information leakage without homologous recognition (\ie, neither homologous $\operatorname{TopK}(\cdot)$ attention nor homologous contrastive loss);
2) adopting exactly the same optimization recipe with MAE.
As demonstrated in~\cref{tab:main_reason}, the model achieves 82.6\% accuracy and 45.9 mIoU, comparably with MAE, suggesting that the information leakage is definitely the culprit here.

\paragraph{Functionality of the $\mathcal{L}_{HomoCon}$.}
To verify that the homologous contrastive loss performs more like a \textit{regularization term} instead of an \textit{individual self-supervision} as in~\cite{ibot,el2021large}, we further pre-train a \textit{MixedAE} with $\mathcal{L}_{HomoCon}$ only 
following the settings in~\cref{sec:ablation}.
As demonstrated in~\cref{tab:loss_function}, \textit{MixedAE} performs well when the reconstruction loss $\mathcal{L}_{recon}$ is utilized only or together with the homologous contrastive loss $\mathcal{L}_{HomoCon}$.
However, when adopting $\mathcal{L}_{HomoCon}$ only, \textit{MixedAE} cannot even achieve reasonable transfer performance, suggesting that $\mathcal{L}_{HomoCon}$ cannot work alone without $\mathcal{L}_{recon}$.

\paragraph{Necessity of mixing.}
To verify to necessity of adopting mixing augmentation in our \textit{MixedAE}, we extend MAE with the homologous contrastive $\mathcal{L}_{HomoCon}$ 
by applying~\cref{equ:homo_con}
to patches across different images in groups of 4 for a fair comparison with \textit{MixedAE}, 
which achieves 81.1\% accuracy and 42.9 mIoU as demonstrated in~\cref{tab:mix_and_segment} (2nd \& 3rd rows),
significantly worse than our default \textit{MixedAE}, revealing the necessity of using mixing augmentation.

\paragraph{Necessity of segment embeddings.}
As shown in~\cref{tab:mix_and_segment} (1st \& 3rd rows), we build a \textit{MixedAE without segment embeddings} and achieve 82.2\% accuracy and 44.9 mIoU, significantly worse than our default \textit{MixedAE}, suggesting the importance of adopting segment embeddings to provide necessary information for homologous recognition.



\begin{table}[t]
\begin{center}
\setlength{\tabcolsep}{1.2mm}
\begin{tabular}{l|cc|c|c|ccc|ccc}
\toprule
\multirow{2}*{Method} & Pre-train & Pre-train & ImageNet & ADE20K & \multicolumn{6}{c}{COCO} \\
\cline{4-11}
 & Epochs & GPU-days & Top-1 Acc. & mIoU & AP$^{bb}$ & AP$^{bb}_{50}$ & AP$^{bb}_{75}$ & AP$^{mk}$ & AP$^{mk}_{50}$ & AP$^{mk}_{75}$ \\
\hline
iBOT~\cite{ibot} & 1000$^*$ & 285.0 & 85.0 & 52.2 & 49.9 & 69.5 & 54.1 & 42.9 & 66.5 & 45.9 \\
MAE~\cite{MAE} & 1600 & 151.1 & 85.9 & 53.6 & 54.0 & 72.6 & 59.0 & 46.3 & 69.6 & 50.4 \\
\hline
\rowcolor{backcolor}
\textbf{MixedAE} & 1500 & 159.7 & 86.0 & 53.8 & 54.5 & 73.4 & 59.3 & 46.9 & 70.4 & 50.9 \\
\rowcolor{backcolor}
\textbf{MixedAE} & 1600 & 170.4 & \textbf{86.2}$^{+0.3}$ & \textbf{54.0}$^{+0.4}$ & \textbf{54.6}$^{+0.6}$ & \textbf{73.5}$^{+0.9}$ & \textbf{59.4}$^{+0.4}$ & \textbf{47.1}$^{+0.8}$ & \textbf{70.7}$^{+1.1}$ & \textbf{51.0}$^{+0.6}$ \\
\bottomrule
\end{tabular}
\end{center}
\vspace{-4mm}
\caption{\textbf{Transfer performance comparison with ViT-Large~\cite{VIT}.}
1) Scalability: our \textit{MixedAE} outperforms MAE consistently even with ViT-Large.
2) Efficiency: \textit{MixedAE} achieves a better trade-off between the computational overhead and the transfer performance regardless of the architecture size.
$^*$: effective epochs following iBOT~\cite{ibot}.
}
\label{tab:vit_large}
\end{table}


\begin{table*}[t]
\vspace{-.2em}
\centering
\subfloat[
    \textbf{Main cause of performance degradation}
    is indeed the information leakage brought by na\"ive mixing without homologous recognition.
    \label{tab:main_reason}
    ]{
    \begin{minipage}{0.29\linewidth}{
        \begin{center}
        \tablestyle{1pt}{1.05}
        \begin{tabular}{x{48}x{40}x{24}x{24}}
        & Masked SA & acc & mIoU \\
        \shline
        MAE & & \textbf{82.7} & \textbf{46.1} \\
        \hline
        Na\"ive &  & 82.4 & 45.0 \\
        Mixing & \checkmark & 82.6 & 45.9 \\
        \end{tabular}
        \end{center}
    }\end{minipage}
}
\hspace{2em}
\subfloat[
    \textbf{Functionality of the $\mathcal{L}_{HomoCon}$.}
        When adopting $\mathcal{L}_{HomoCon}$ alone, \textit{MixedAE} cannot even achieve reasonable transfer performance.
    \label{tab:loss_function}
    ]{
    \centering
    \begin{minipage}{0.29\linewidth}{
        \begin{center}
        \tablestyle{1pt}{1.05}
        \begin{tabular}{x{40}x{40}x{24}x{24}}
            $\mathcal{L}_{recon}$ & $\mathcal{L}_{HomoCon}$ & acc & mIoU \\
            \shline
            \checkmark & & 82.4 & 45.0 \\
            & \checkmark & 7.8 & 8.3 \\
            \checkmark & \checkmark & \baseline{\textbf{82.7}} & \baseline{\textbf{46.4}} \\
        \end{tabular}
        \end{center}
    }
    \end{minipage}
}
\hspace{2em}
\subfloat[
    \textbf{Necessity of the mixing and segment embeddings.}
        The best transfer performance is achieved when both are adopted.
    \label{tab:mix_and_segment}
    ]{
    \centering
    \begin{minipage}{0.29\linewidth}{
        \begin{center}
        \tablestyle{1pt}{1.05}
        \begin{tabular}{x{40}x{40}x{24}x{24}}
            Mixing & SE & acc & mIoU \\
            \shline
            \checkmark & & 82.2 & 44.9 \\
            & \checkmark & 81.1 & 42.9 \\
            \checkmark & \checkmark & \baseline{\textbf{82.7}} & \baseline{\textbf{46.4}} \\
        \end{tabular}
        \end{center}
    }
    \end{minipage}
}
\vspace{-1.em}
\caption{\textbf{More \textit{MixedAE} ablation experiments} with ViT-B/16.
Default settings are marked in \colorbox{baselinecolor}{gray}.}
\label{tab:app_exp}
\vspace{-.5em}
\end{table*}


\section{More Analysis}\label{app:analysis}

\paragraph{Exploration for other augmentations.}
As discussed in~\cref{sec:baseline}, 
based on mixing, we observe that common augmentation strategies will increase mutual information (MI) between the model input and the reconstruction target,
suggesting that data augmentations like random augmentation and color jittering might \textit{not be suitable} or \textit{require specific designs} for MIM, which will be explored in the future work.

\paragraph{Limitations.}
Although demonstrating significant performance, we notice that the $\operatorname{TopK}(\cdot)$ sampling accuracy in homologous attention still cannot achieve 100\% as shown 
in~\cref{fig:accuracy},
for which there exist several potential reasons accounting: 
1) The background patches might be included during random cropping inevitably, which are difficult for attention-based methods to recognize.
2) There is still further improvement space for \textit{MixedAE}. For example, more strict verification than homologous contrastive loss is an appealing future work direction.


\section{More Visualizations}\label{app:visualization}

\begin{figure*}[!t]
  \centering
  \begin{subfigure}{1.0\linewidth}
    \includegraphics[width=1.0\linewidth]{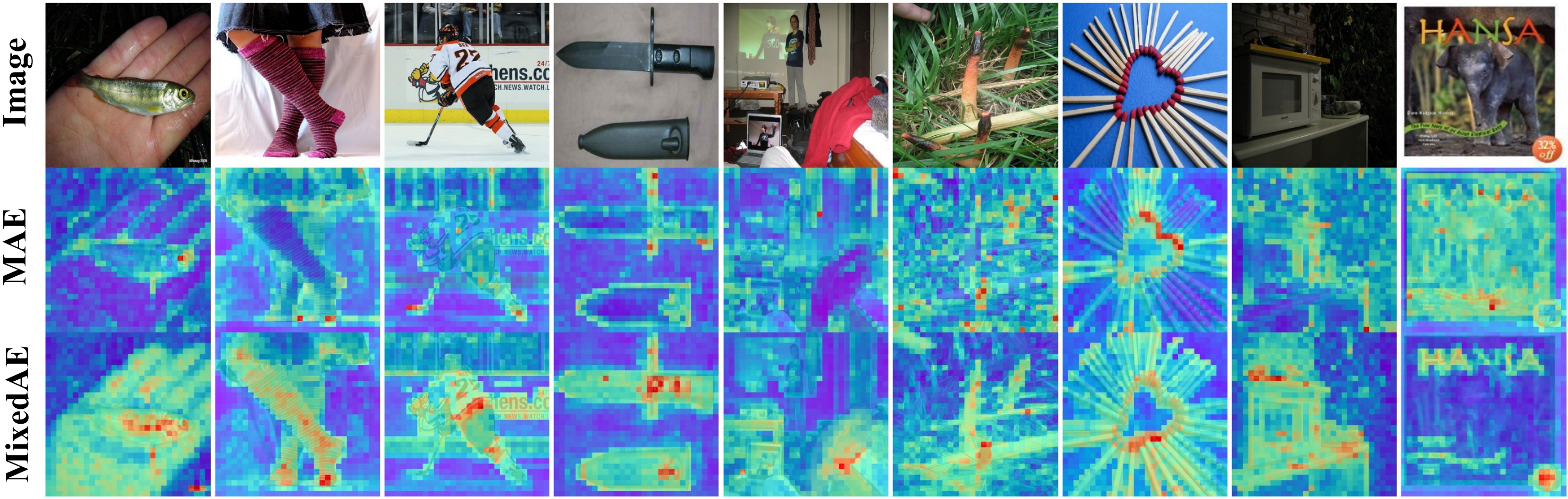}
  \end{subfigure}
  \vspace{-1mm}
  \begin{subfigure}{1.0\linewidth}
    \includegraphics[width=1.0\linewidth]{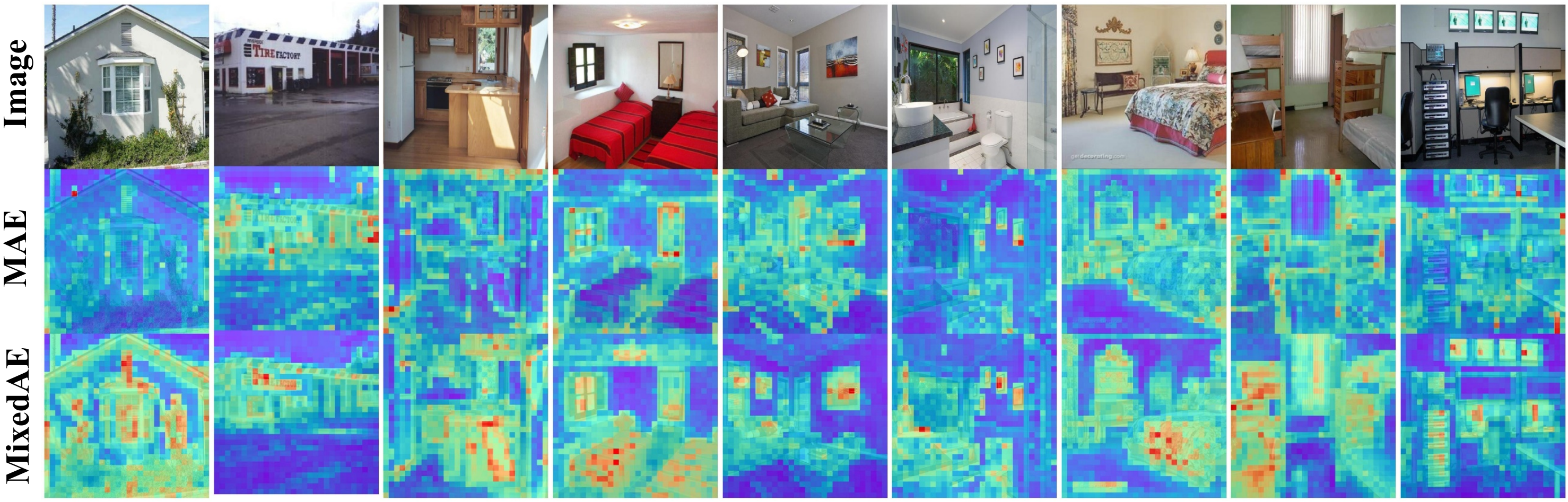}
  \end{subfigure}
  \vspace{-1mm}
  \begin{subfigure}{1.0\linewidth}
    \includegraphics[width=1.0\linewidth]{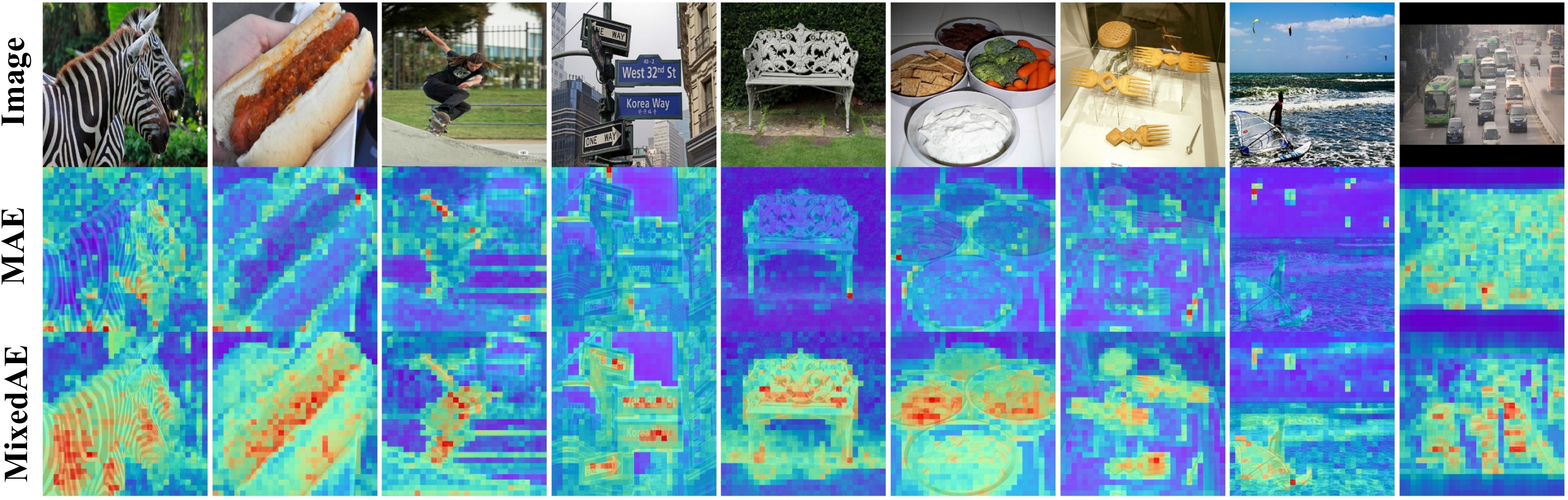}
  \end{subfigure}
  \vspace{-1mm}
  \begin{subfigure}{1.0\linewidth}
    \includegraphics[width=1.0\linewidth]{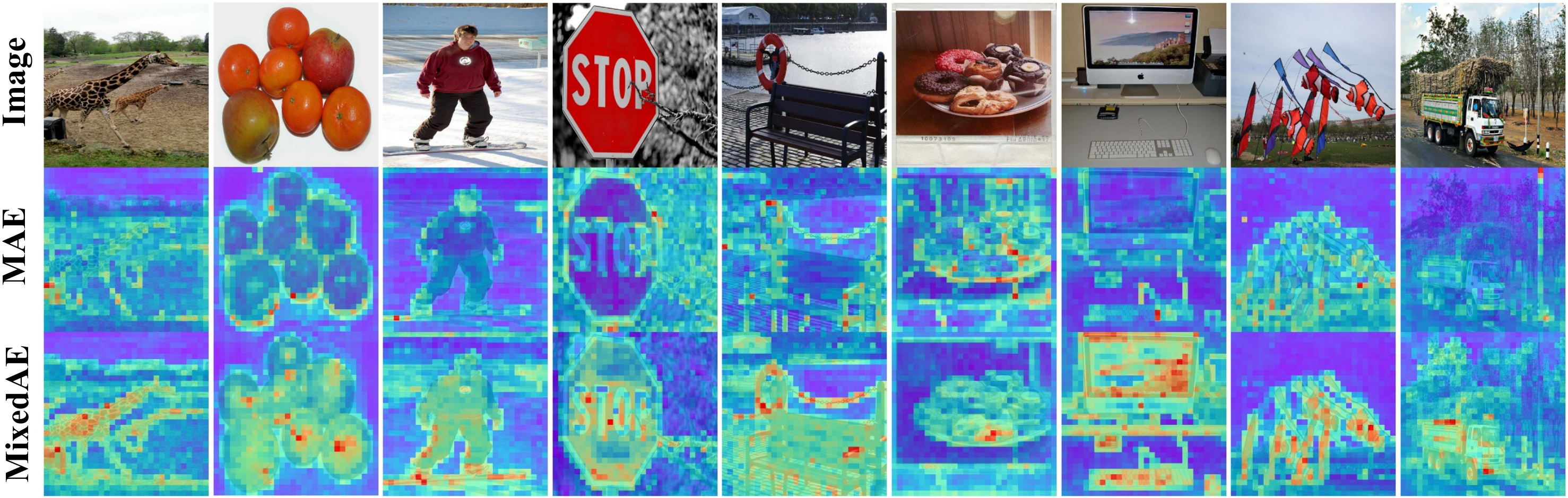}
  \end{subfigure}
  \vspace{-6mm}
  \caption{\textbf{More visualizations of attention maps} from ImageNet-1K (1-3 rows), ADE20K (4-6 rows) and COCO (7-12 rows).
  }
  \label{fig:more_att_maps}
\end{figure*}

We provide more visualizations of the attention mapes learnt by MAE and \textit{MixedAE} on images from ImageNet-1K~\cite{deng2009imagenet}, ADE20K~\cite{ADE20K} and Microsoft COCO~\cite{lin2014microsoft} datasets in~\cref{fig:more_att_maps}.
As demonstrated, our \textit{MixedAE} can generate more reasonable and discriminative attention maps, revealing the effectiveness of \textit{MixedAE} to conduct object-aware pre-training without any specifically designed modules (\eg, K-means~\cite{chen2021multisiam} and object discovery network~\cite{henaff2022object}).

\end{document}